\documentclass{article}
\usepackage{spconf,amsmath,graphicx}
\usepackage{colortbl, multirow, makecell, booktabs, bm, amssymb, enumitem}
\usepackage{gensymb}
\usepackage{flushend}
\definecolor{red}{rgb}{1, 0.7, 0.7}
\definecolor{orange}{rgb}{1, 0.85, 0.7}
\definecolor{yellow}{rgb}{1.0, 1.0, 0.7}

\title{Gaussian Splatting with Reflections Guided by What is Seen}
\name{Yiming Liang$^{\star}$\qquad Tianyu Xiao$^{\star}$ \qquad Hiroshi Ishikawa\thanks{$^\star$ denotes equal contribution. This work was partially supported by JSPS KAKENHI Grant Number 25K22823. \\ Copyright 2026 IEEE. Published in 2026 IEEE International Conference on Image Processing (ICIP), scheduled for 13-17 September 2026 in Tampere, Finland. Personal use of this material is permitted. Permission from IEEE must be obtained for all other uses, in any current or future media, including reprinting/republishing this material for advertising or promotional purposes, creating new collective works, for resale or redistribution to servers or lists, or reuse of any copyrighted component of this work in other works.}}
\address{Department of Computer Science and Communications Engineering, Waseda University, Japan}
\begin{document}
%
\maketitle
\begin{abstract}
While novel view synthesis methods like Neural Radiance Fields (NeRF) and 3D Gaussian Splatting (3DGS) have shown impressive performance on common scenes, they often struggle with capturing view-dependent specular effects, particularly near-field reflections. Existing methods utilize environment maps, which only represent lights from infinity, or environment Gaussians, which rely solely on indirect supervision from reflections, are insufficient for accurately modeling such effects. Here, we propose a framework based on 2D Gaussian Splatting (2DGS) that leverages directly observable environment information in input images. Our approach also integrates a hybrid reflection appearance representation and view-dependent specular modeling. Experiments on public benchmarks and a newly curated near-field reflection dataset demonstrate that our method achieves superior rendering quality and geometric accuracy.
\end{abstract}
\section{Introduction}
\label{sec:intro}
Novel view synthesis (NVS) aims to generate photorealistic images of a scene from unseen viewpoints using a set of captured images. Neural Radiance Fields (NeRF)~\cite{nerf} have significantly advanced the field by leveraging a multi-layer perceptron (MLP) as an implicit representation of the radiance field. More recently, 3D Gaussian Splatting (3DGS)~\cite{3dgs} has shifted the paradigm back toward explicit representations. 3DGS adopts anisotropic 3D Gaussians as primitives and enables real-time rendering via efficient rasterization without sacrificing quality. Despite its rapid progress, modeling view-dependent effects, particularly high-frequency specular reflections, remains an open challenge for 3DGS-based NVS.

The original 3DGS framework does not explicitly model specular reflections. Instead, it utilizes Spherical Harmonics (SH) to model view-dependent appearance, which can only simulate reflections to a limited degree. This representation treats reflections as an anisotropic property of each surface point rather than modeling the physics of light transport. Consequently, reflections lack real supervision via reflected light rays, and observations from different input views cannot be properly correlated. This often results in blurred reflections or geometry distortions that arise to compensate for appearance modeling, particularly for near-field reflections. Recent works~\cite{gaussianshader, 3dgsdr, refgs} address this by introducing environment maps, but these assume light originates from infinity and thus fail to capture the spatially varying reflections from near-field objects in real-world scenes. While EnvGS~\cite{envgs} attempts to bridge this gap by replacing the environment map with environment Gaussians, it is optimized purely through indirect reflection supervision, leading to inaccurate environment geometry and degraded reflection quality.

\begin{figure*}[t]
\centering
\setlength{\tabcolsep}{2pt} 
\scalebox{0.97}{
    \includegraphics[width=0.95\linewidth]{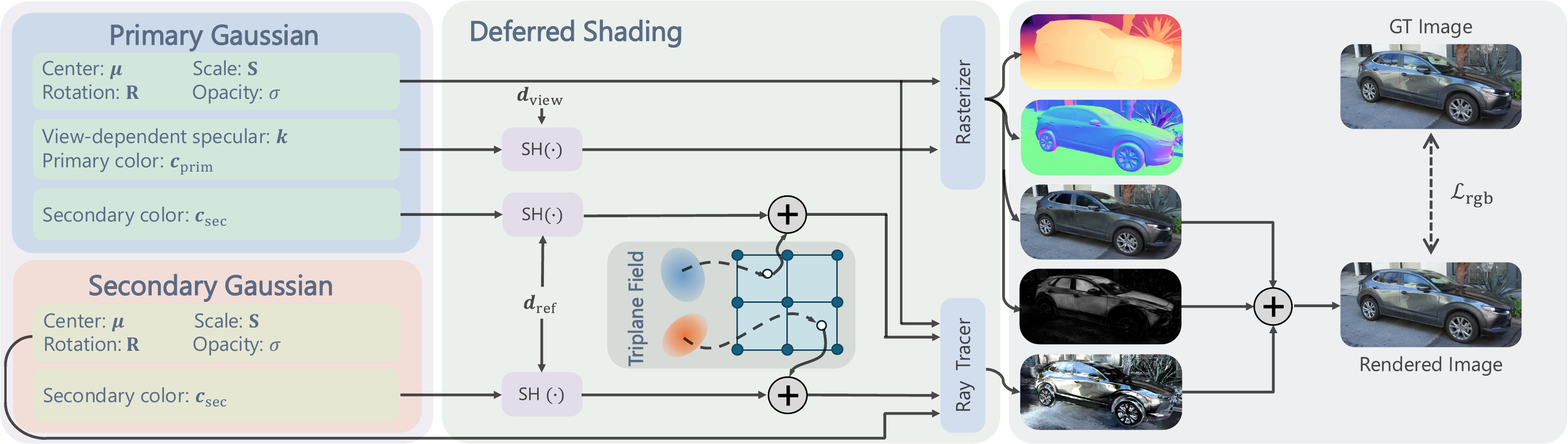}
}
\caption{Overview of our method. We use two Gaussian sets: primary for the base scene and secondary for near-field environment light. Reflections are rendered by ray-tracing through both sets, with colors decoded from a hybrid triplane and SH coefficients. Final pixel color is obtained by blending diffuse and reflection components using a view-dependent specular weight.}
\label{fig:overview}
\end{figure*}

In this work, we propose a framework based on 2D Gaussian Splatting (2DGS) \cite{2dgs} to achieve high-quality NVS in scenes with near-field reflections. Our key insight is that the sources of the near-field environment light reflected by surfaces are often directly observable from other viewpoints in the input images. Leveraging this observation, we represent the scene using two sets of Gaussians: primary Gaussians for the base scene (objects directly observable from input images) and secondary Gaussians for nearby objects appearing as reflections. Rendering consists of a diffuse pass and a reflection pass. In the diffuse pass, the diffuse component is rendered from the primary Gaussians via rasterization. In the reflection pass, secondary rays are traced from the surfaces of the base scene, and intersected with the union of the primary and secondary Gaussians. Since the geometry of the primary Gaussians is directly constrained by the input images, using them in the reflection pass ensures that the environment light originates from accurate 3D positions. To enhance rendering quality, we adopt a hybrid representation that leverages both triplane-based color fields and SH in the reflection pass. We also assign a separate SH to each primary Gaussian in the reflection pass to avoid conflicts with the diffuse appearance modeling. Additionally, we incorporate a data-driven, view-dependent specular weight model using another set of SH to capture the complex lighting variations in real-world scenes.

We evaluate our method on multiple widely used benchmarks and a new custom real-world dataset, NearRef, which consists of four indoor scenes with prominent near-field reflections. Quantitative and qualitative results demonstrate that our approach produces sharper reflections, more accurate geometry, and better disentanglement than prior methods, while reconstructing more faithful environment lighting. 

Our contributions can be summarized as follows: 
\begin{itemize}[itemsep=1pt]
   \item We propose a 2DGS-based framework for NVS with near-field reflections that leverages direct observation for environment light modeling.
   \item We introduce a hybrid appearance model for the reflection pass and a view-dependent specular model to enhance the quality.
   \item We achieve state-of-the-art results on both public benchmarks and the new NearRef dataset featuring near-field reflections. The code and dataset will be released publicly.  
\end{itemize}
\section{Related Work}
\label{sec:related}
\subsection{Novel View Synthesis}
Neural Radiance Fields (NeRF)~\cite{nerf} implicitly encode a scene into an MLP, which is optimized via differentiable volume rendering. Many subsequent methods have been proposed to further push the boundaries of the implicit representation paradigm~\cite{barron2021mip, barron2023zip, muller2022instant}. Despite these advances, NeRF-based methods remain computationally expensive. 
Recently, 3D Gaussian Splatting (3DGS)~\cite{3dgs} has introduced an explicit representation using anisotropic 3D Gaussians, enabling real-time rendering via efficient rasterization, and has seen improvements and extensions in various domains~\cite{2dgs, lu2024scaffold, Wang2025HyRFHR, liang2025himor}.
However, accurately modeling high-frequency specular reflections remains an open challenge.

\subsection{Reflective Scene Reconstruction}
Capturing specular reflections with high fidelity has been explored within both NeRF- and 3DGS-based frameworks. Ref-NeRF~\cite{refnerf} improves specular modeling in NeRF by introducing reflection-direction encoding based on SH, while subsequent works~\cite{yang2024spec, nde} explore alternative representations. Beyond directional encoding, NeRF-Casting~\cite{nerfcasting} models specular components by explicitly casting secondary reflection rays. 3DGS-based approaches often employ environment maps to model reflections~\cite{gaussianshader, 3dgsdr, refgs}, which are queried using the reflection direction and jointly optimized with the Gaussian representation. 3DGS-DR~\cite{3dgsdr} and Ref-GS~\cite{refgs} adopt deferred shading for more stable normal estimation, with Ref-GS~\cite{refgs} further introducing a neural environment map. However, their assumption that all environment lights originate at infinity makes them unsuitable for modeling spatially varying near-field reflections in real-world scenes. To mitigate this, EnvGS~\cite{envgs} employs a separate set of environment Gaussians as a 3D representation of environment light. However, its optimization purely through indirect reflection supervision without direct multi-view constraints often leads to degraded quality and inaccurate environment geometry.
\section{Method}
\label{sec:methods}
\begin{table*}[ht]
\centering
\renewcommand{\arraystretch}{0.5} 
\caption{Quantitative NVS results on existing datasets~\cite{nerfcasting, envgs, refnerf}. \colorbox{red}{Best}, \colorbox{orange}{second best}, and \colorbox{yellow}{third best} results are highlighted.}
\scalebox{0.88}{ 
\begin{tabular}{l | ccc | ccc | ccc | ccc}
    \toprule
    & \multicolumn{3}{c|}{NeRF-Casting~\cite{nerfcasting}} & \multicolumn{3}{c|}{EnvGS~\cite{envgs}} & \multicolumn{3}{c|}{Ref-Real~\cite{refnerf}} & \multicolumn{3}{c}{\textbf{Average}} \\
    Method & PSNR$\uparrow$ & SSIM$\uparrow$ & LPIPS$\downarrow$ & PSNR$\uparrow$ & SSIM$\uparrow$ & LPIPS$\downarrow$ & PSNR$\uparrow$ & SSIM$\uparrow$ & LPIPS$\downarrow$ & PSNR$\uparrow$ & SSIM$\uparrow$ & LPIPS$\downarrow$ \\
    \midrule
    3DGS~\cite{3dgs} & \cellcolor{yellow}29.7161 & \cellcolor{red}0.8930 & \cellcolor{red}0.1442 & \cellcolor{yellow}26.6340 & \cellcolor{orange}0.8831 & \cellcolor{yellow}0.1525 & \cellcolor{yellow}24.0680 & 0.6600 & \cellcolor{yellow}0.2532 & \cellcolor{yellow}27.1485 & \cellcolor{red}0.8131 & \cellcolor{orange}0.1824 \\
    2DGS~\cite{2dgs} & 28.7545 & 0.8728 & 0.1825 & 26.2465 & 0.8752 & 0.1738 & 23.8770 & \cellcolor{orange}0.6699 & 0.2749 & 26.5714 & 0.8057 & 0.2114 \\
    GShader~\cite{gaussianshader} & 27.1522 & 0.8453 & 0.2243 & 25.4491 & 0.8553 & 0.1982 & 23.1039 & 0.6370 & 0.3098 & 25.4247 & 0.7781 & 0.2470 \\
    3DGS-DR~\cite{3dgsdr} & 28.8993 & \cellcolor{orange}0.8805 & 0.1956 & 25.8913 & 0.8594 & 0.1908 & 22.8938 & 0.6303 & 0.2913 & 26.2290 & 0.7924 & 0.2264 \\
    RefGS~\cite{refgs} & 24.2534 & 0.8122 & 0.2408 & 25.6502 & 0.8536 & 0.2014 & 23.8390 & 0.6493 & 0.2898 & 24.4257 & 0.7671 & 0.2484 \\
    EnvGS~\cite{envgs} & \cellcolor{orange}30.3324 & 0.8734 & \cellcolor{yellow}0.1562 & \cellcolor{orange}27.2666 & \cellcolor{yellow}0.8809 & \cellcolor{orange}0.1478 & \cellcolor{red}24.5972 & \cellcolor{red}0.6706 & \cellcolor{red}0.2438 & \cellcolor{orange}27.7394 & \cellcolor{yellow}0.8075 & \cellcolor{yellow}0.1834 \\
    \textbf{Ours} & \cellcolor{red}30.6185 & \cellcolor{yellow}0.8762 & \cellcolor{orange}0.1523 & \cellcolor{red}27.4223 & \cellcolor{red}0.8835 & \cellcolor{red}0.1442 & \cellcolor{orange}24.4144 & \cellcolor{yellow}0.6684 & \cellcolor{orange}0.2466 & \cellcolor{red}27.8402 & \cellcolor{orange}0.8085 & \cellcolor{red}0.1819 \\
    \bottomrule
\end{tabular}}
\label{tab:existing_dataset_nvs}
\end{table*}

\begin{table*}[ht]
\centering
\caption{Quantitative NVS results on our custom NearRef dataset. \colorbox{red}{Best}, \colorbox{orange}{second best}, and \colorbox{yellow}{third best} results are highlighted.}
\renewcommand{\arraystretch}{0.9}
\scalebox{0.72}{ 
\begin{tabular}{l | ccc | ccc | ccc | ccc | ccc}
    \toprule
    & \multicolumn{3}{c|}{Black Monitor} & \multicolumn{3}{c|}{White Monitor} & \multicolumn{3}{c|}{Coffee Machine} & \multicolumn{3}{c|}{Cleaner} & \multicolumn{3}{c}{\textbf{Average}} \\
    Method & PSNR$\uparrow$ & SSIM$\uparrow$ & LPIPS$\downarrow$ & PSNR$\uparrow$ & SSIM$\uparrow$ & LPIPS$\downarrow$ & PSNR$\uparrow$ & SSIM$\uparrow$ & LPIPS$\downarrow$ & PSNR$\uparrow$ & SSIM$\uparrow$ & LPIPS$\downarrow$ & PSNR$\uparrow$ & SSIM$\uparrow$ & LPIPS$\downarrow$ \\
    \midrule
    3DGS~\cite{3dgs} & \cellcolor{yellow}28.4841 & \cellcolor{yellow}0.9140 & \cellcolor{yellow}0.1071 & \cellcolor{yellow}35.0724 & \cellcolor{red}0.9698 & \cellcolor{yellow}0.0938 & \cellcolor{yellow}29.5640 & \cellcolor{orange}0.9414 & \cellcolor{orange}0.0983 & \cellcolor{yellow}33.1301 & \cellcolor{yellow}0.9598 & \cellcolor{yellow}0.0882 & \cellcolor{yellow}31.5626 & \cellcolor{yellow}0.9463 & \cellcolor{yellow}0.0969 \\
    2DGS~\cite{2dgs} & 27.5291 & 0.9035 & 0.1181 & 33.5976 & 0.9647 & 0.1153 & 28.5130 & 0.9287 & 0.1270 & 31.8991 & 0.9536 & 0.1070 & 30.3847 & 0.9376 & 0.1168 \\
    GShader~\cite{gaussianshader} & 24.1317 & 0.8507 & 0.1817 & 26.9088 & 0.9143 & 0.1956 & 24.6470 & 0.8701 & 0.2018 & 28.3498 & 0.9292 & 0.1537 & 26.0093 & 0.8911 & 0.1832 \\
    3DGS-DR~\cite{3dgsdr} & 27.1847 & 0.8992 & 0.1254 & 33.2582 & 0.9623 & 0.1194 & 28.8980 & 0.9315 & 0.1216 & 28.2506 & 0.9313 & 0.1358 & 29.3979 & 0.9310 & 0.1256 \\
    RefGS~\cite{refgs} & 27.8328 & 0.8970 & 0.1382 & 32.8170 & 0.9572 & 0.1271 & 28.7207 & 0.9186 & 0.1479 & 25.5781 & 0.9129 & 0.1702 & 28.7372 & 0.9214 & 0.1458 \\
    EnvGS~\cite{envgs} & \cellcolor{orange}29.8510 & \cellcolor{orange}0.9237 & \cellcolor{orange}0.0917 & \cellcolor{orange}35.2431 & \cellcolor{yellow}0.9684 & \cellcolor{orange}0.0856 & \cellcolor{orange}30.4561 & \cellcolor{yellow}0.9402 & \cellcolor{yellow}0.1014 & \cellcolor{orange}33.6592 & \cellcolor{orange}0.9607 & \cellcolor{orange}0.0830 & \cellcolor{orange}32.3023 & \cellcolor{orange}0.9483 & \cellcolor{orange}0.0904 \\
    \textbf{Ours} & \cellcolor{red}30.8711 & \cellcolor{red}0.9372 & \cellcolor{red}0.0730 & \cellcolor{red}35.5418 & \cellcolor{orange}0.9696 & \cellcolor{red}0.0742 & \cellcolor{red}30.9279 & \cellcolor{red}0.9429 & \cellcolor{red}0.0965 & \cellcolor{red}33.9113 & \cellcolor{red}0.9607 & \cellcolor{red}0.0797 & \cellcolor{red}32.8130 & \cellcolor{red}0.9526 & \cellcolor{red}0.0809 \\
    \bottomrule
\end{tabular}}
\label{tab:custom_dataset_nvs}
\end{table*}

\def\Gprim{\mathcal{G}_{\text{prim}}}
\def\Gsec{\mathcal{G}_{\text{sec}}}
\def\mui{\bm{\mu}_i}
\def\Si{\mathbf{S}_i}
\def\Ri{\mathbf{R}_i}
\def\sigmai{\bm{\sigma}_i}
\def\ki{\bm{k}_i}
\def\ci{\bm{c}_i}
\def\cpi{\bm{c}_{\text{prim},i}}
\def\csi{\bm{c}_{\text{sec},i}}
\def\cp{\bm{c}_{\text{prim}}}
\def\Cp{\bm{C}_{\text{prim}}}
\def\Cpi{\bm{C}_{\text{prim},i}}
\def\cs{\bm{c}_{\text{sec}}}
\def\k{\bm{k}}
\def\Ct{\bm{C}_{\text{tri}}}
\def\dv{\bm{d}_\text{view}}
\def\dr{\bm{d}_\text{ref}}

\subsection{Preliminary}
Our method is built upon 2D Gaussian Splatting (2DGS)~\cite{2dgs}, which utilizes 2D Gaussians as primitives to represent scenes. Compared to 3D Gaussians, 2D Gaussians provide superior geometric consistency and well-defined surface normals. Each 2D Gaussian is specified by its center $\bm{\mu}$ in 3D space, scale and rotation matrices $\mathbf{S}=\text{diag}(s_x, s_y, 0)$ and $\mathbf{R}=(\bm{t}_x,\bm{t}_y, \bm{0})$, opacity $\sigma$, and an SH coefficient vector $\bm{c}$, parameterizing view-dependent color. The normal vector can be computed as: $\bm{n} = \bm{t}_x \times \bm{t}_y$. 
To render a set $\mathcal{G}$ of 2D Gaussians into an image, consider the ray going from the point of view through each pixel.
Let $g_i=(\mui,\Si,\Ri,\sigmai,\ci)$ be the $i$th Gaussian (in the order from the nearest to the camera to the farthest) intersecting the ray, with $g_i(\cdot)$ denoting its evaluation at the intersection.
Then, the color $\bm{C}$ at the pixel is given by:
\begin{equation}
    \bm{C}=\sum_{i=1}^{N}T_i\alpha_i\bm{C}_i,\,\alpha_i=\sigma_i\cdot g_i(\cdot),\,T_i=\prod_{j=1}^{i-1}(1 - \alpha_j),
\end{equation}
The color $\bm{C}_i$ at $g_i$ is $\bm{C}_i=\mathbf{SH}(\bm{c}_i,\dv)$. $\text{SH}(\bm{x},\dv)$ denotes the SH with coefficient vector $\bm{x}$ and view direction $\dv$; in bold letters $\mathbf{SH}$ signifies an RGB vector.
Other values such as depth and normal can be rendered similarly by replacing $\bm{C}_i$ with the depth $d_i$ or the normal $\bm{n}_i$ at each Gaussian. This rendering process can be implemented either through tile-based rasterization~\cite{2dgs} or ray-tracing~\cite{envgs}: we denote by $\text{Rasterizer}(\bm{v}|\,\mathcal{G})$ or $\text{RayTracer}(\bm{v}|\,\mathcal{G})$ the rendering of a value $\bm{v}$ with respect to a Gaussian set $\mathcal{G}$ by the two methods.

\subsection{Scene Representation}
We represent the scene using two distinct sets of 2D Gaussians: a primary Gaussian $\Gprim$, and a secondary Gaussian $\Gsec$. Specifically, we define these sets as follows:
\begin{equation}
\begin{aligned}
\Gprim&=\{(\mui, \Si, \Ri, \sigmai, \cpi, \csi, \ki)\}_{i=1}^N,  \\
\Gsec&=\{(\mui, \Si, \Ri, \sigmai,\csi)\}_{i=N+1}^M. \\
\end{aligned}
\end{equation}
The primary Gaussian $\Gprim$ has two color coefficient vectors $\cpi$ and $\csi$ for the diffuse and reflection passes, respectively. It also has SH coefficients $\ki$ for view-dependent specular weight that controls the blending between diffuse and specular components (see Sec.~\ref{sec:view-dependent specular}).

The rendering process adopts a deferred shading \cite{3dgsdr, refgs, envgs}. In the diffuse pass, we rasterize $\Gprim$ to obtain the diffuse color $\bm{C}_{\text{diff}}$, normal $\bm{N}$, depth $D$, and specular weight $K$:
\begin{equation}
\begin{aligned}
(\bm{C}_{\text{diff}},\bm{N},D,&K)=\\
&\text{Rasterizer}(\cdot|\,\Gprim)*(\Cp,\bm{n},d,k),
\end{aligned}
\end{equation}
where $*$ signifies a component-wise evaluation and
$\Cpi=\mathbf{SH}(\cpi,\dv), k_i=\text{SH}(\ki,\dv)$.

In the reflection pass, we compute the specular component $\bm{C}_{\text{ref}}$. Using rendered depth $D$, the intersection point between the primary ray and the object surface can be determined as:$\,\bm{p}=\bm{o} + D\cdot\bm{d}_{\text{view}}$, where $\bm{o}$ is the camera origin. The secondary ray, $\bm{r}(t) = \bm{p} + t\cdot \dr$, induced by reflection, is cast along the reflection direction $\dr=\bm{d}_{\text{view}} - 2(\bm{N}\cdot \bm{d}_{\text{view}})\bm{N}$. The specular component $\bm{C}_{\text{ref}}$ is then obtained by ray-tracing through the union of both sets of Gaussians:
\begin{equation}
\bm{C}_{\text{ref}}=\text{RayTracer}(\Ct+\mathbf{SH}(\cs,\dr)|\,\Gprim\cup\Gsec),
\label{eq:raytracing}
\end{equation}
where $\Ct$ is decoded from a Triplane (Sec. \ref{sec:hybrid}). Note that we employ $\cs$ instead of $\cp$ for $\Gprim$ to avoid conflicts with the diffuse pass.
Finally, the pixel color is blended as: 
\begin{equation}
\label{eq:blending}
\bm{C}=(1 - K)\cdot\bm{C}_{\text{diff}} + K\cdot\bm{C}_{\text{ref}}.
\end{equation}
\textbf{Discussion.} A key contribution of our approach is the use of $\Gprim\cup\Gsec$ during the reflection rendering.
Prior methods rely solely on a representation of environment light optimized only through an indirect supervision from reflection, lacking multi-view constraints. This leads to geometric ambiguity and loss of details. By incorporating $\Gprim$, which benefits from direct multi-view supervision via primary rays, we leverage its high-fidelity geometry in the reflection rendering. This ensures that directly observable regions are rendered with high quality in reflection, while $\Gsec$ fills in the remaining environment that is not directly visible in the input images.

\begin{figure*}[t]
\centering
\setlength{\tabcolsep}{2pt}
\scalebox{0.98}{
    \begin{tabular}{@{}cccccc@{}}
        \rotatebox{90}{\hspace{5pt}\textbf{Hatchback}} &
        \includegraphics[  width=0.19\linewidth, trim={0 0 0 60}, clip]{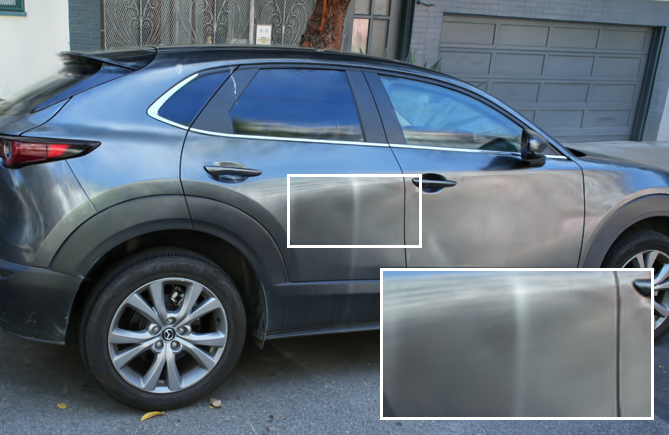} &
        \includegraphics[  width=0.19\linewidth, trim={0 0 0 60}, clip]{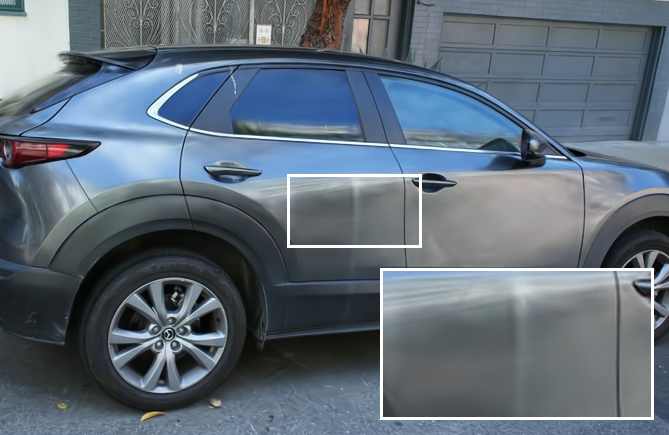} & 
        \includegraphics[  width=0.19\linewidth, trim={0 0 0 60}, clip]{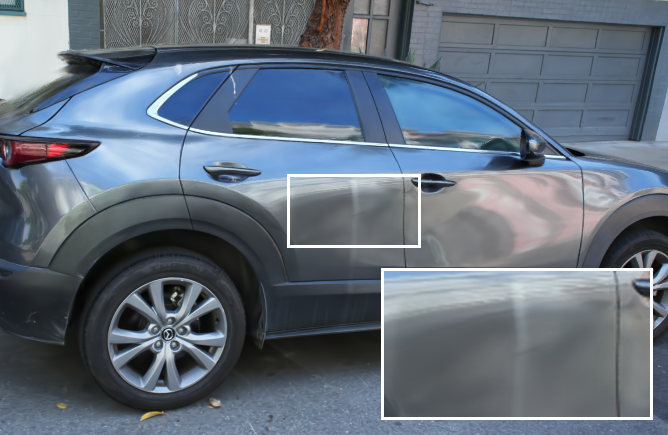} & 
        \includegraphics[  width=0.19\linewidth, trim={0 0 0 60}, clip]{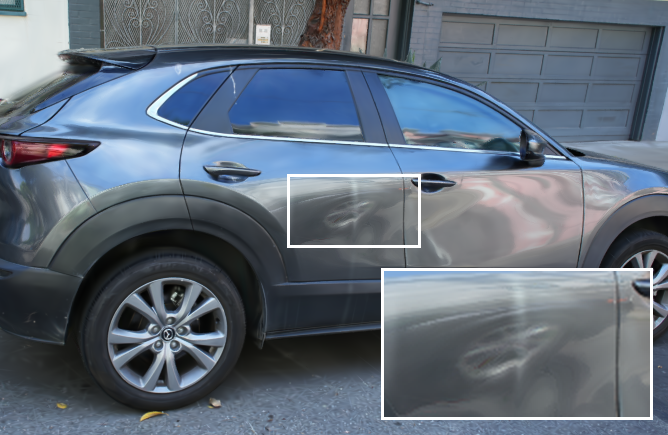} & 
        \includegraphics[  width=0.19\linewidth, trim={0 0 0 60}, clip]{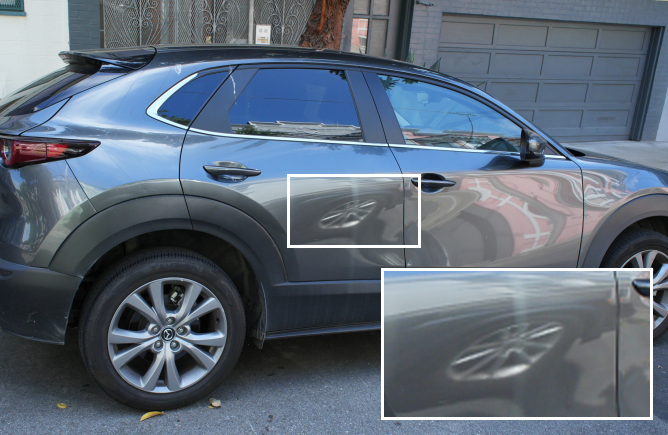} \\ 
        \rotatebox{90}{\hspace{18pt} \textbf{Audi}} &
        \includegraphics[width=0.19\linewidth, trim={0 0 0 10}, clip]{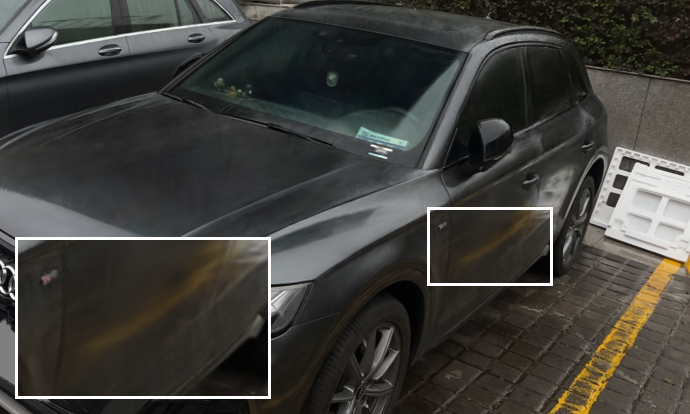} &
        \includegraphics[width=0.19\linewidth, trim={0 0 0 10}, clip]{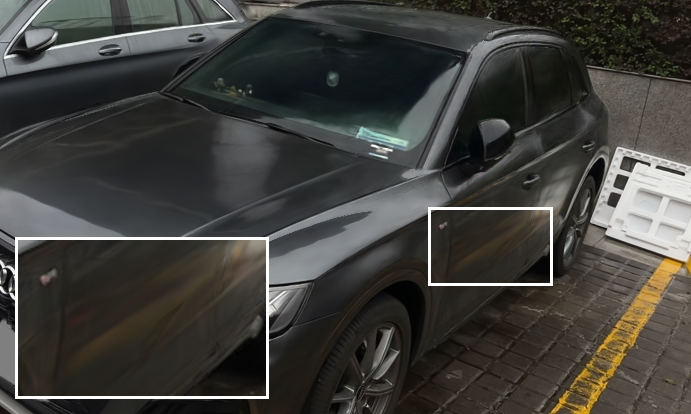} &
        \includegraphics[width=0.19\linewidth, trim={0 0 0 10}, clip]{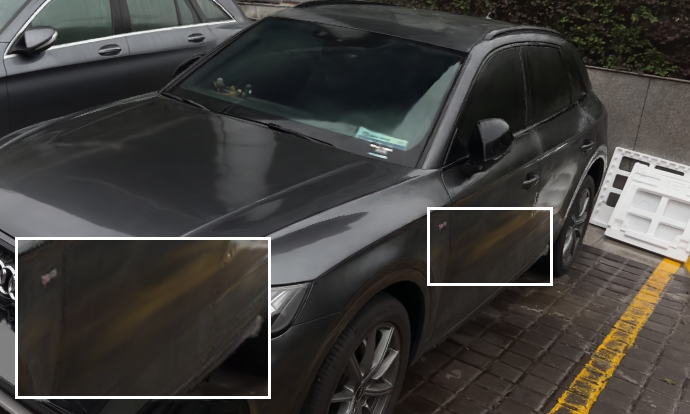} & 
        \includegraphics[width=0.19\linewidth, trim={0 0 0 10}, clip]{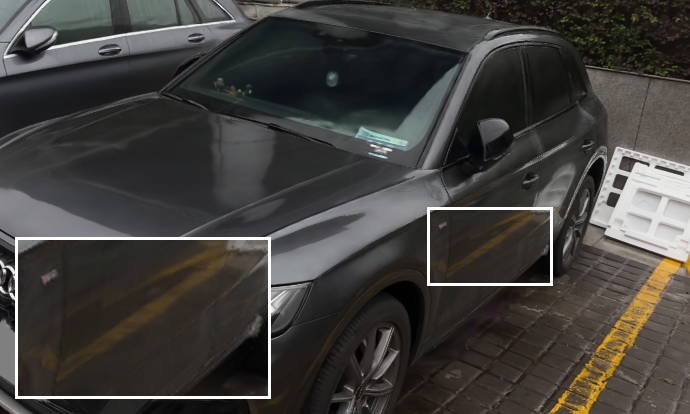} & 
        \includegraphics[width=0.19\linewidth, trim={0 0 0 10}, clip]{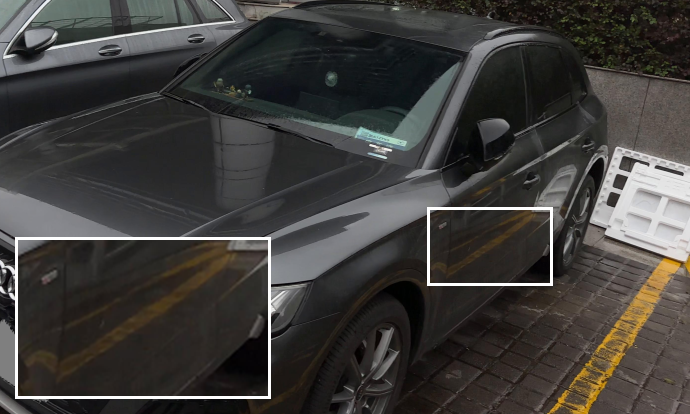} \\
        
        & 3DGS & 3DGS-DR & EnvGS & Ours & GT \\
    \end{tabular}
}
\caption{Qualitative NVS comparison on the existing datasets~\cite{nerfcasting, envgs}.}
\label{fig:existing_dataset_nvs_visual}
\end{figure*}

\begin{figure}[t]
\centering
\setlength{\tabcolsep}{1pt}
\renewcommand{\arraystretch}{0.5}
\scalebox{0.9}{
    \begin{tabular}{@{}cccccc@{}}
        \rotatebox{90}{\hspace{2pt} \textbf{Black Monitor}} &
        \includegraphics[width=0.19\linewidth,trim={0 0 0 30}, clip]{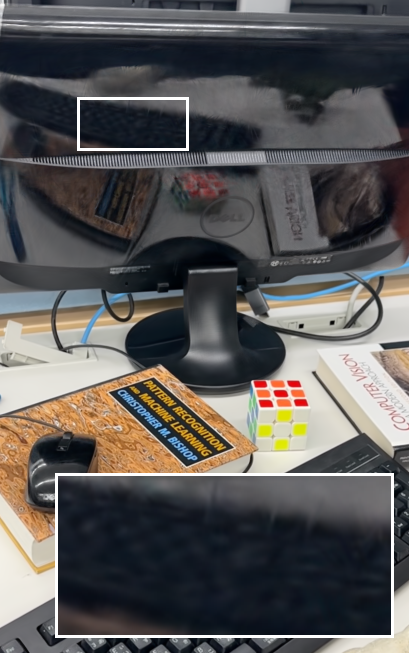} &
        \includegraphics[width=0.19\linewidth,trim={0 0 0 30}, clip]{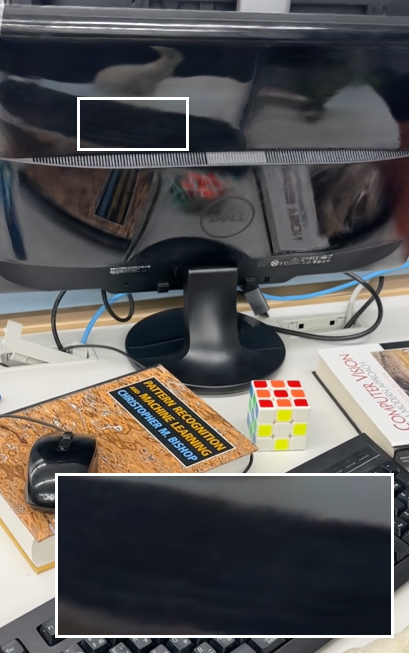} & 
        \includegraphics[width=0.19\linewidth,trim={0 0 0 30}, clip]{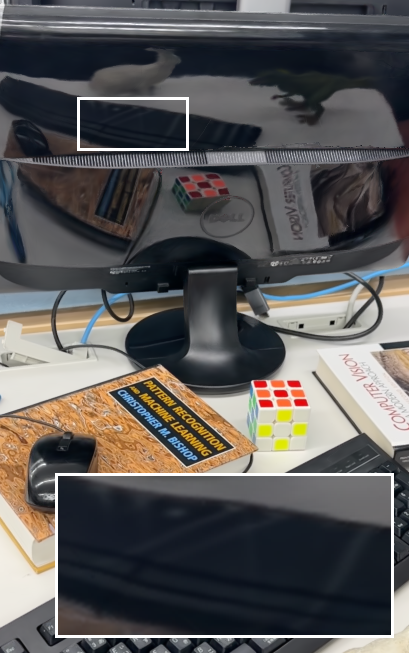} & 
        \includegraphics[width=0.19\linewidth,trim={0 0 0 30}, clip]{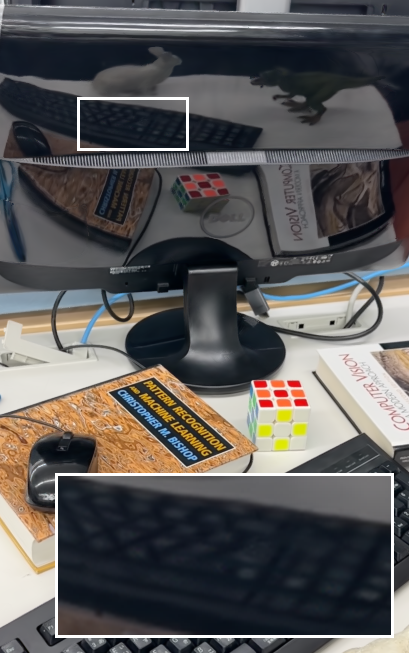} & 
        \includegraphics[width=0.19\linewidth,trim={0 0 0 30}, clip]{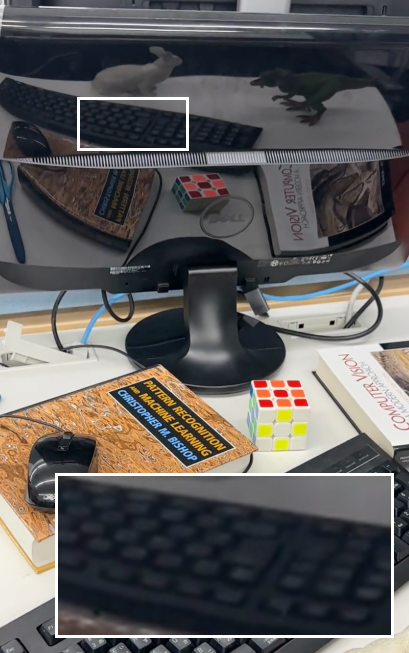} \\ 
        \rotatebox{90}{\hspace{13pt} \textbf{Cleaner}} &
        \includegraphics[width=0.19\linewidth,trim={0 0 0 50}, clip]{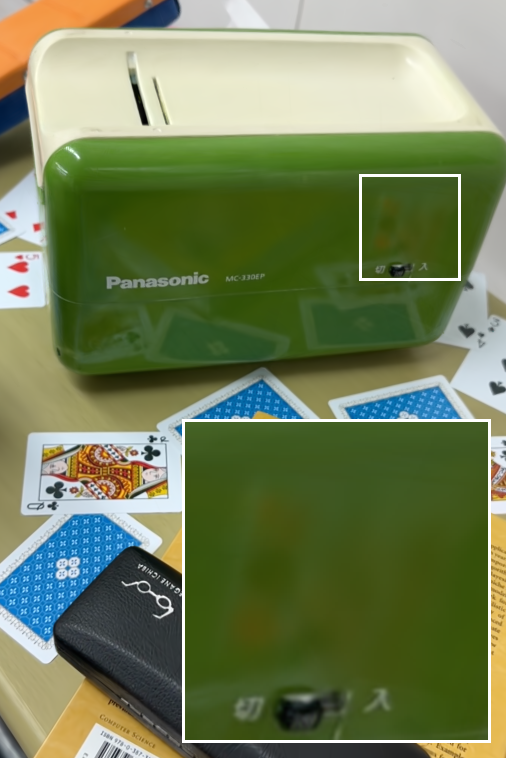} &
        \includegraphics[width=0.19\linewidth,trim={0 0 0 50}, clip]{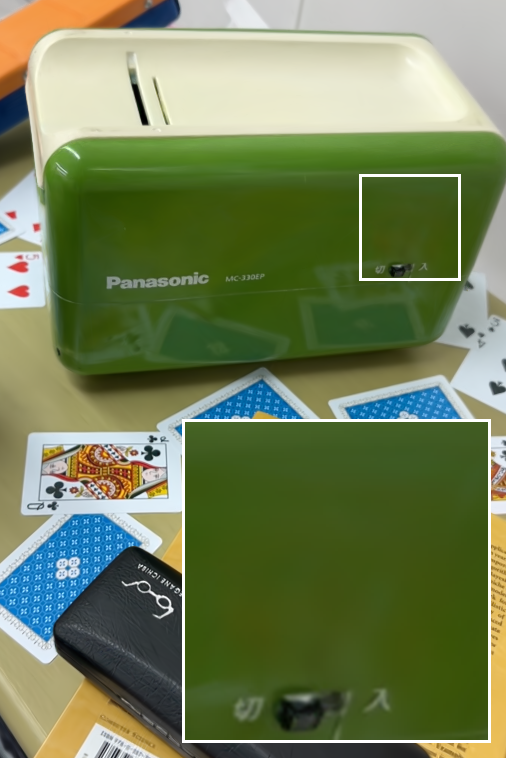} &
        \includegraphics[width=0.19\linewidth,trim={0 0 0 50}, clip]{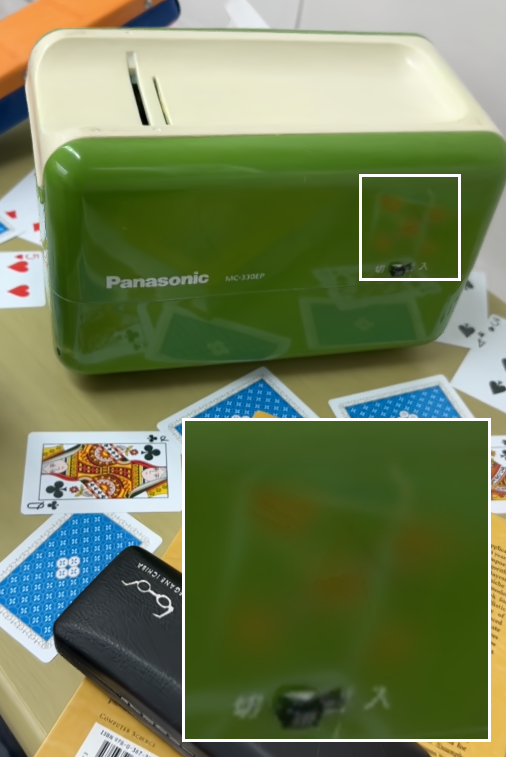} & 
        \includegraphics[width=0.19\linewidth,trim={0 0 0 50}, clip]{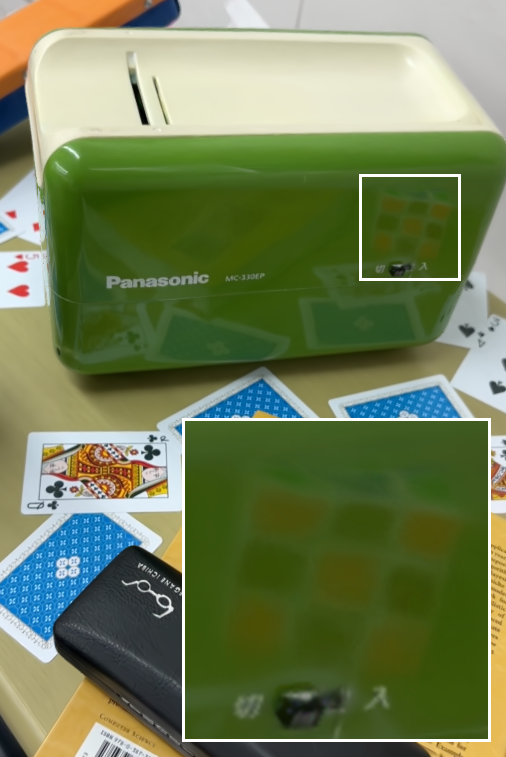} & 
        \includegraphics[width=0.19\linewidth,trim={0 0 0 50}, clip]{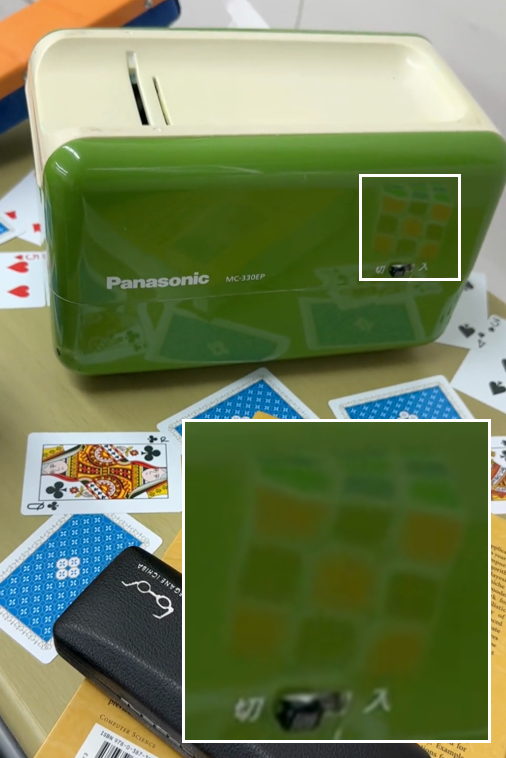} \\

        & 3DGS & 3DGS-DR & EnvGS & Ours & GT \\
    \end{tabular}
}
\caption{Qualitative NVS comparison on our custom NearRef dataset.}
\label{fig:custom_dataset_nvs_visual}
\end{figure}

\subsection{Hybrid Appearance Model}
\label{sec:hybrid}
We introduce a hybrid color representation for both the primary and secondary Gaussians when rendered in reflection pass. The hybrid color is defined in \eqref{eq:raytracing}, where $\Ct$ is a color decoded from a Triplane representation. The Triplane encodes the 3D scenes into three axis-aligned orthogonal $H\times W$ feature planes $\mathbf{F}_{\text{xy}}, \mathbf{F}_{\text{yz}}, \mathbf{F}_{\text{zx}}\in\mathbb{R}^{H\times W\times L}$ of $L$-dimensional feature vectors. For a Gaussian center $\bm{\mu}$ normalized to $[0,\,1]^3$ as \cite{mipnerf360}, features are sampled via bilinear interpolation from these planes and decoded into RGB color via an MLP ($\mathbb{R}^{3\cdot L}\to$ RGB) with one hidden layer as:
\begin{equation}
\Ct = \text{MLP}([\mathbf{F}_{\text{xy}}(\mu_x, \mu_y),\,\mathbf{F}_{\text{yz}}(\mu_y, \mu_z),\,\mathbf{F}_{\text{zx}}(\mu_z, \mu_x)]).
\end{equation}
This neural prior provides a spatial smoothness constraint that regularizes the optimization process, preventing estimated surface normals from converging to noisy local minima. Simultaneously, high-frequency view-dependency is captured by the view-dependent SH coefficients $\bm{c}_{\text{sec}}$.

\subsection{View-dependent Specular}
\label{sec:view-dependent specular}
We model the specular weight $K$ using SH to account for the Fresnel effect, where reflectivity increases drastically as the viewing direction approaches a grazing angle. While the Schlick approximation~\cite{schlick} is commonly used in physics-based rendering, it is highly sensitive to noisy normal estimates, which can destabilize optimization in this case. By employing a data-driven SH representation, the weight is optimized directly from the input images. This allows the model to flexibly capture the view-dependency of specular weight and approximate ambient occlusion not explicitly considered by our simplified blending framework as in Eq. \eqref{eq:blending}.

\subsection{Losses}
The overall loss function is defined as:
\begin{equation}
\mathcal{L}=\lambda_{\text{rgb}}\mathcal{L}_{\text{rgb}} + \lambda_{\text{consist}}\mathcal{L}_{\text{consist}} + \lambda_{\text{mono}}\mathcal{L}_{\text{mono}},
\end{equation}
where the image reconstruction loss $\mathcal{L}_{\text{rgb}}$ is a weighted sum of $\mathcal{L}_1$ loss, D-SSIM~\cite{SSIM} loss, and LPIPS~\cite{lpips} loss. $\mathcal{L}_{\text{consist}}$ is the normal consistency loss, which aligns the rendered normal with the gradients of the depth map. Finally, $\mathcal{L}_{\text{mono}}$ supervises the rendered normal using estimates from a pretrained model~\cite{wang2025moge}.

\section{Experiments}
\label{sec:experiments}
\begin{figure}[t]
  \centering
  \setlength{\tabcolsep}{1pt}
  \scalebox{1}{
    \begin{tabular}{@{}ccc@{}}
      \includegraphics[width=0.32\linewidth, trim = 200 10 60 120, 
        clip]{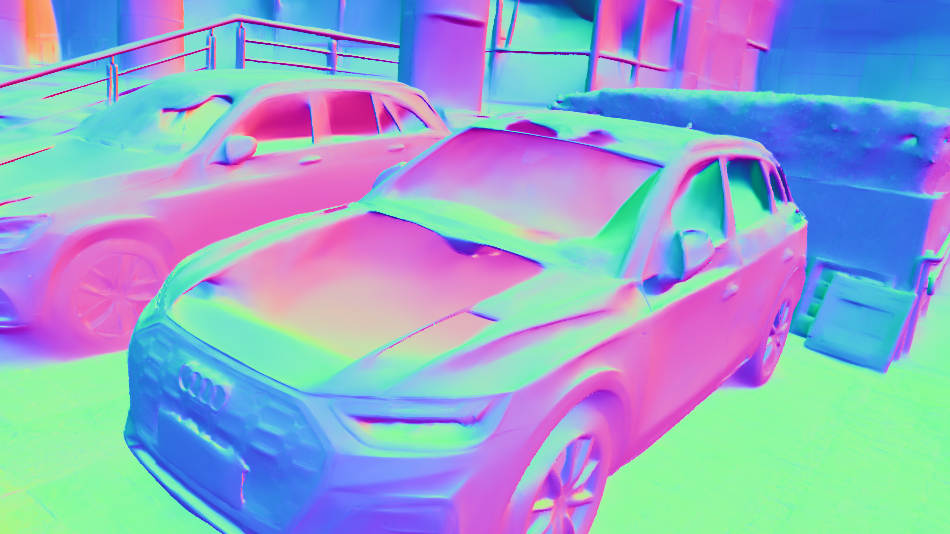} 
      &
      \includegraphics[width=0.32\linewidth, trim = 200 10 60 120, 
        clip]{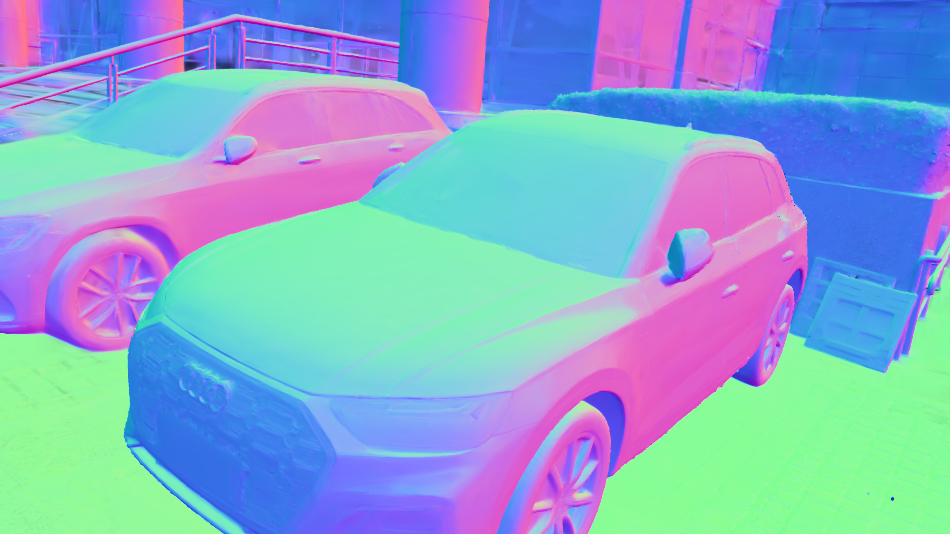}
      &
      \includegraphics[width=0.32\linewidth, trim = 200 10 60 120, 
        clip]{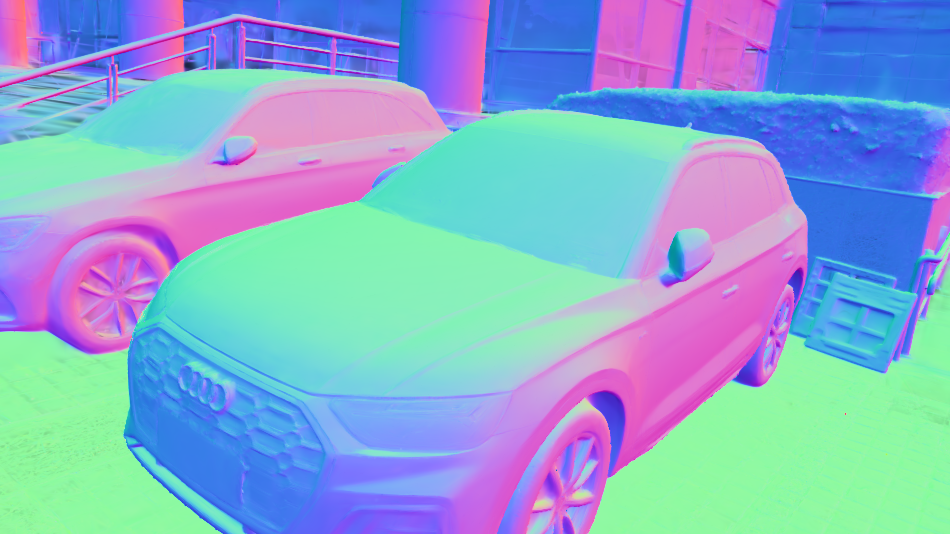}\\
      2DGS &EnvGS & Ours \\
  \end{tabular}}
  \caption{Qualitative comparison of surface normal estimation.}
  \label{fig:normal}
\end{figure}
\noindent{\textbf{Implementation details.}}
Primary Gaussians $\mathcal{G}_{\text{prim}}$ are initialized from COLMAP~\cite{colmap} point clouds, while secondary Gaussians $\mathcal{G}_{\text{sec}}$ are initialized by placing five Gaussians per cell in a $32^3$ uniform grid spanning the scene. We employ a progressive optimization strategy: first reconstructing with $\mathcal{G}_{\text{prim}}$ alone, then introducing reflection rendering via $\mathcal{G}_{\text{sec}}$ for coarse reflections, and finally jointly refining $\mathcal{G}_{\text{prim}}\cup\mathcal{G}_{\text{sec}}$ to capture near-field details. We adopt the normal propagation and color sabotage from~\cite{3dgsdr} to aid optimization. For reflection ray-tracing, we use the differentiable 2D Gaussian ray-tracer from EnvGS~\cite{envgs}. The optimization takes approximately 3-6 hours on an NVIDIA RTX 4090 GPU with peak memory usage under 12 GB. During inference, the rendering speed is around 10 FPS.

\noindent{\textbf{Datasets.}}
We evaluate our method on several existing datasets for reflective scenes: NeRF-Casting~\cite{nerfcasting}, EnvGS~\cite{envgs}, and Ref-Real~\cite{refnerf}. To address the lack of datasets containing near-field reflections, we collected a custom dataset, NearRef, comprising four indoor scenes. These scenes were captured using an iPhone 14 Pro. In each scene, a reflective object is placed at the center of the scene with additional objects nearby to create high-frequency near-field reflections.  

\noindent{\textbf{Baselines.}}
We evaluate our approach against various baselines categorized by their representations for reflection modeling. We compare against general-purpose NVS methods, 3DGS~\cite{3dgs} and 2DGS~\cite{2dgs}, which rely on SH. We also include GShader~\cite{gaussianshader}, 3DGS-DR~\cite{3dgsdr}, and Ref-GS~\cite{refgs}, which are based on environment maps. For 3D environment modeling, we compare against EnvGS~\cite{envgs}.

\begin{figure}[t]
\centering
\setlength{\tabcolsep}{1pt}
\renewcommand{\arraystretch}{0.5}
\scalebox{0.94}{
    \begin{tabular}{@{}cccc@{}} 
      \rotatebox{90}{\hspace{12pt}\textbf{EnvGS}} &
      \includegraphics[width=0.32\linewidth, trim = 210 110 90 120, 
        clip]{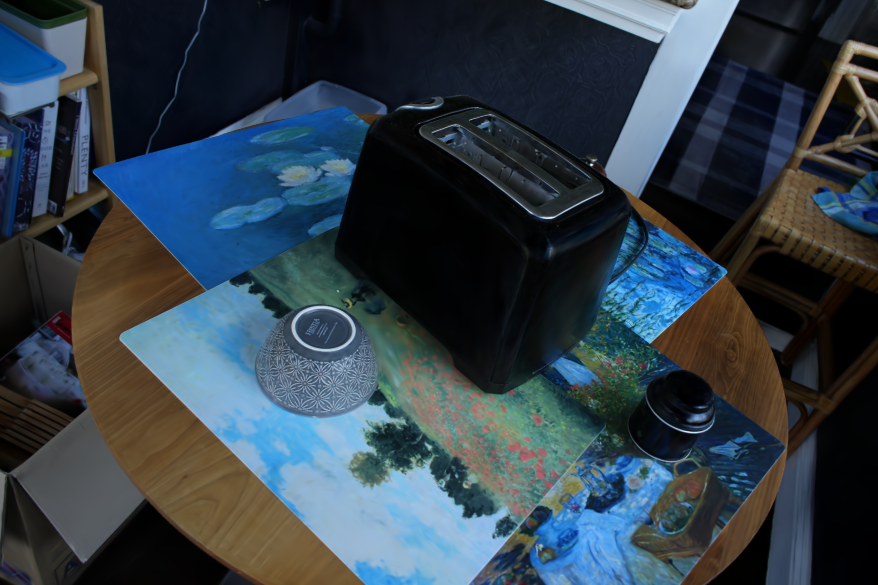} &
      \includegraphics[width=0.32\linewidth,trim = 210 110 90 120, 
        clip ]{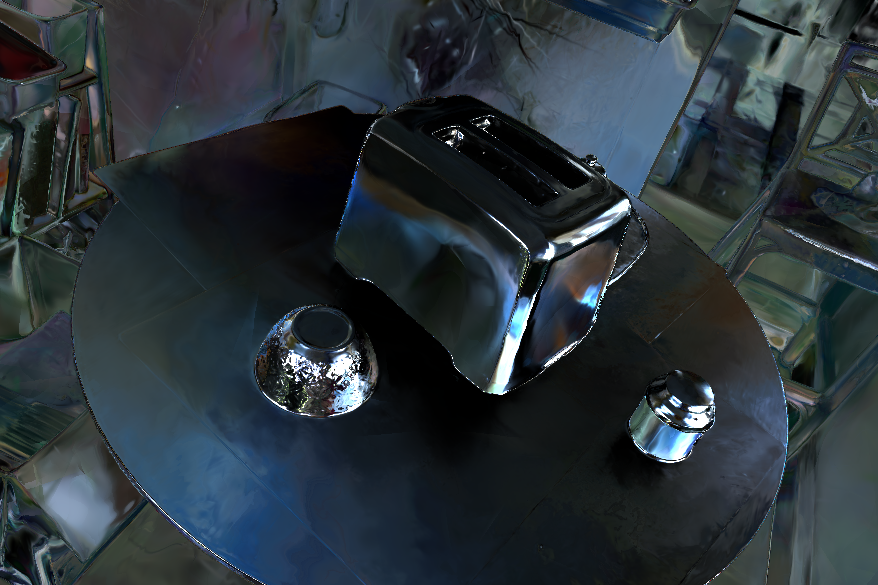} &
      \includegraphics[width=0.32\linewidth,trim = 210 110 90 120, 
        clip ]{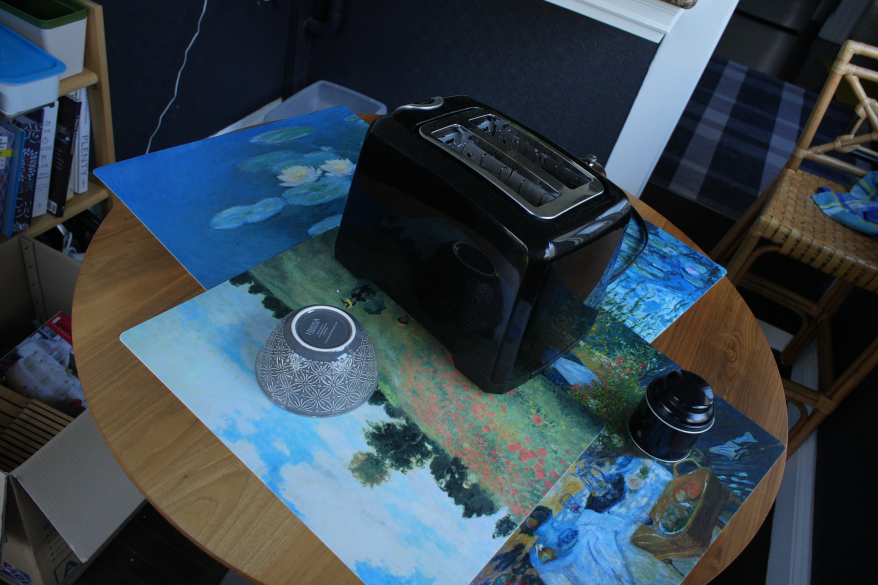}  \\
      \rotatebox{90}{\hspace{20pt}\textbf{Ours}} &
      \includegraphics[width=0.32\linewidth,trim = 210 110 90 120, 
        clip]{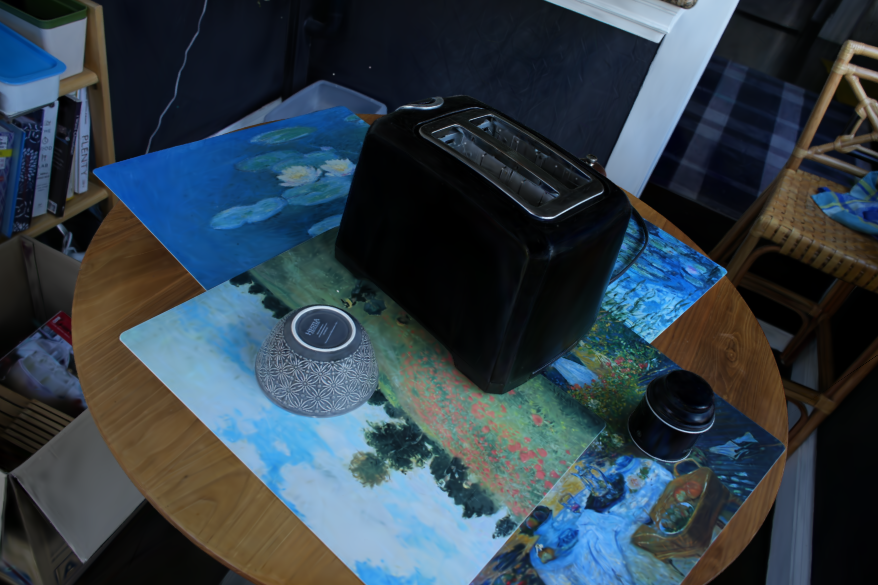} &
      \includegraphics[width=0.32\linewidth, trim = 210 110 90 120, 
        clip]{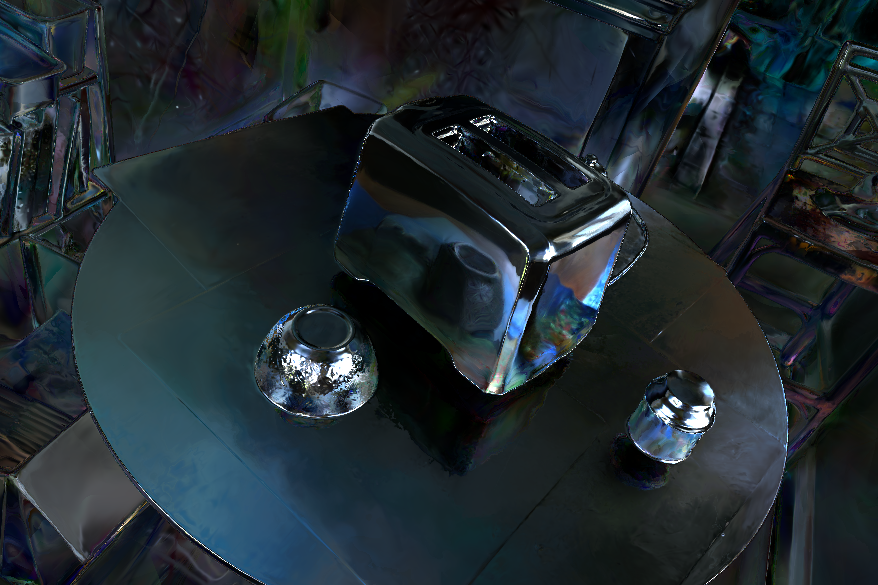} &
      \includegraphics[width=0.32\linewidth, trim = 210 110 90 120, 
        clip]{images/decompose/rendered/0232_gt.png}  \\
      & Diffuse & Reflection & GT \\
  \end{tabular}
}
\caption{Disentanglement of diffuse and reflection components.}
\label{fig:decomposition}
\end{figure}
\begin{figure}[t]
  \centering
  \setlength{\tabcolsep}{1pt}
  \scalebox{1}{
    \begin{tabular}{@{}cc@{}}
      \includegraphics[width=0.48\linewidth]{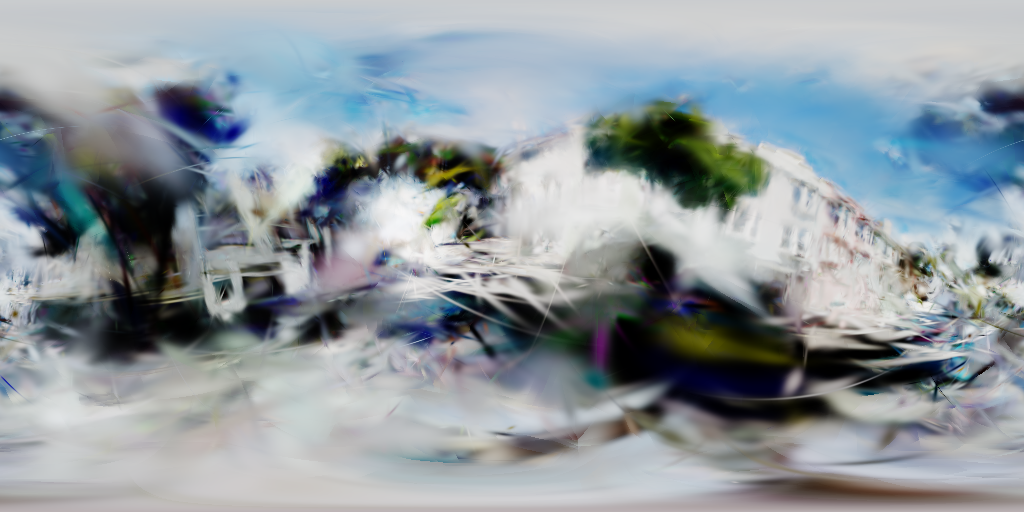} 
      &
      \includegraphics[width=0.48\linewidth]{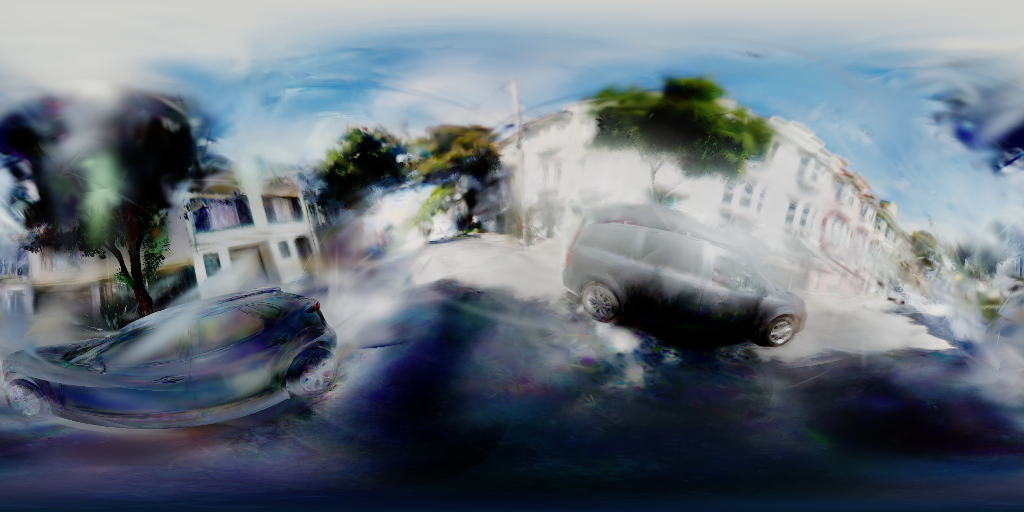}\\
      EnvGS & Ours \\
  \end{tabular}}
  \caption{Reconstructed environment light.}
  \label{fig:envmap}
\end{figure}

\subsection{Baseline Comparisons}

\noindent{\textbf{Novel view synthesis.}}
As shown in Tab.~\ref{tab:existing_dataset_nvs} and Tab.~\ref{tab:custom_dataset_nvs}, our method achieves competitive performance on existing datasets and demonstrates superiority on our custom NearRef dataset. Qualitative results in Fig.~\ref{fig:existing_dataset_nvs_visual} and Fig.~\ref{fig:custom_dataset_nvs_visual} show that 3DGS~\cite{3dgs} and 3DGS-DR~\cite{3dgsdr} produce blurry or even no reflections due to the limited capacity of SH and the violation of infinity assumptions. While EnvGS~\cite{envgs} also employs Gaussians to model reflection, our method achieves significantly better recovery of details. This shows that our method successfully leverages direct observation by integrating primary Gaussians $\mathcal{G}_{\text{prim}}$ into the reflection rendering, leading to a substantial improvement in the near-field reflection fidelity. More results are shown in the supplementary material.

\noindent{\textbf{Normal estimation.}}
We compare our reconstructed surface normals against those of 2DGS~\cite{2dgs} and EnvGS~\cite{envgs} in Fig.~\ref{fig:normal}. For 2DGS, the normals in reflective regions collapse inward as the model overfits view-dependent effects, indicating the limited capacity of SH. Compared to EnvGS, our method produces normals with finer details. 

\noindent{\textbf{Decomposition.}}
As shown in Fig.~\ref{fig:decomposition}, our method achieves superior disentanglement of diffuse and reflection components compared to EnvGS~\cite{envgs}. While EnvGS suffers from blurry reflections, our framework recovers significantly sharper reflections with high-frequency details. 

\noindent{\textbf{Environment light.}}
We visualize the reconstructed environment light as an environment map in Fig.~\ref{fig:envmap}. Our method is  significantly more faithful to the underlying geometry and achieves higher visual fidelity compared to EnvGS~\cite{envgs}. Notably, EnvGS fails to produce some reflected environment (e.g., black vehicle), resulting in missing details as shown in Fig.~\ref{fig:existing_dataset_nvs_visual}. This highlights the necessity of leveraging direct observation, as indirect reflection alone cannot provide sufficient constraints.

\begin{table}[t]
  \centering
  \caption{Ablation studies. ``\checkmark'' denotes included components for variant A--F. Best results in \textbf{bold}.}
  \scalebox{0.65}{
  \begin{tabular}{l|ccccc|ccc}
    \toprule
    Methods &$\mathcal{G}_{\text{prim}} \cup \mathcal{G}_{\text{sec}}$ & $\bm{c}_{\text{ref}} \neq \bm{c}_{\text{prim}}$ & $\bm{c}_{\text{sec}}$ &$\bm{c}_{\text{tri}}$ & $\bm{k}$ &PSNR$\uparrow$ & SSIM$\uparrow$ & LPIPS$\downarrow$ \\
    \midrule
     A&&&  &  & & 27.4097& 0.8437 & 0.1827 \\
     B&\checkmark& & &  & &27.4368& 0.8443& 0.1799 \\
     C&\checkmark&\checkmark &\checkmark & & &27.5775 & 0.8462 &0.1766 \\
     D&\checkmark&\checkmark  && \checkmark && 27.6532& 0.8470 &0.1791  \\
     E&\checkmark&\checkmark  &\checkmark &\checkmark & &27.6678 &0.8479 & 0.1772 \\
     F (Full)&\checkmark&\checkmark & \checkmark & \checkmark & \checkmark & \textbf{27.7905} & \textbf{0.8497} & \textbf{0.1752} \\
    
    \bottomrule 
  \end{tabular}
  }
  \label{tab:ablation_study}
\end{table}

\subsection{Ablation Studies}
In Tab.~\ref{tab:ablation_study}, we show the results of ablation studies using the Hatchback scene in the NeRF-Casting dataset~\cite{nerfcasting}. \textbf{(B)} Naively using $\mathcal{G}_{\text{prim}}\cup\mathcal{G}_{\text{sec}}$ with $\bm{c}_{\text{ref}}= \bm{c}_{\text{prim}}$ in reflection rendering yields limited gains due to interference during optimization. \textbf{(C)} Introducing a separate secondary color $\bm{c}_{\text{sec}}$ for primary Gaussians provides a large improvement. \textbf{(D)} Replacing $\bm{c}_{\text{sec}}$ with triplane-based $\bm{c}_{\text{tri}}$ improves PSNR and SSIM by enhancing spatial smoothness, but results in higher LPIPS, indicating the lack of high-frequency detail. \textbf{(E)} Combining $\bm{c}_{\text{tri}}$ and SH-based $\bm{c}_{\text{sec}}$ produces better overall performance, validating our hybrid appearance design. \textbf{(F)} With further incorporation of view-dependent specular $\bm{k}$, our full model achieves the best performance across all metrics.
\section{Conclusion}
\label{sec:conclusion}
We proposed a 2DGS-based framework for novel view synthesis in reflective scenes, specifically targeting near-field reflections. By integrating primary Gaussians into the reflection rendering, our method leverages directly observable scene information to provide geometric constraints for environment light reconstruction. Combined with a hybrid appearance model and a data-driven, view-dependent specular model, our approach outperforms prior methods, yielding superior rendering quality with sharper reflections. Future research directions include developing a more efficient ray-tracer, modeling surface roughness, and extending the framework to account for refraction.
%

\bibliographystyle{IEEEbib}
\bibliography{strings,refs}

\raggedend
\appendix
\begin{center}
	{\large\bf Supplementary Material}
\end{center}
\section{Additional results}
\begin{figure*}[t]
\centering
\setlength{\tabcolsep}{2pt}
\scalebox{0.97}{
    \begin{tabular}{@{}cccccc@{}}
        \rotatebox{90}{\hspace{10pt} \textbf{Compact}} &
        \includegraphics[width=0.19\linewidth]{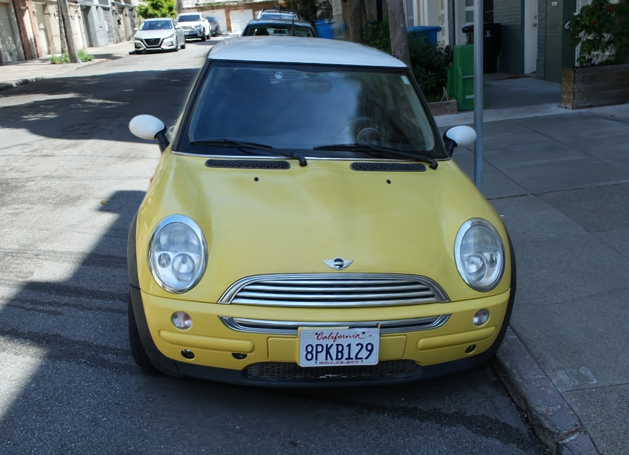} &
        \includegraphics[width=0.19\linewidth]{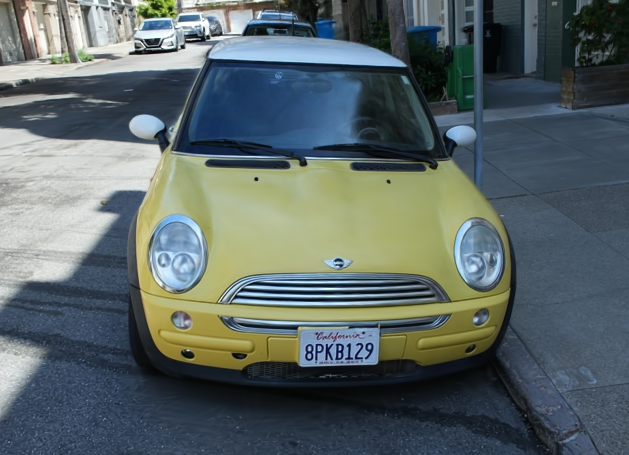} & 
        \includegraphics[width=0.19\linewidth]{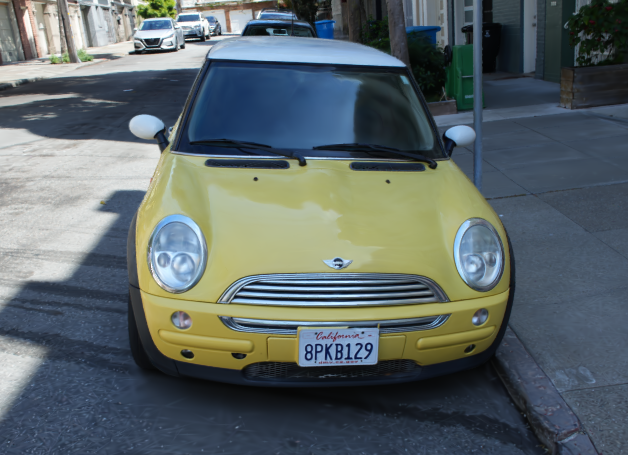} & 
        \includegraphics[width=0.19\linewidth]{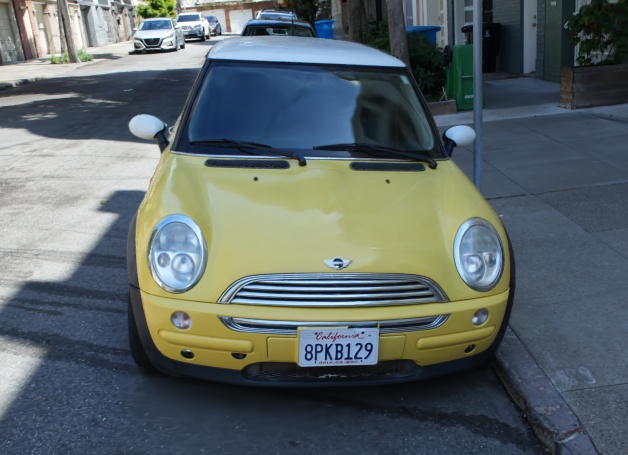} & 
        \includegraphics[width=0.19\linewidth]{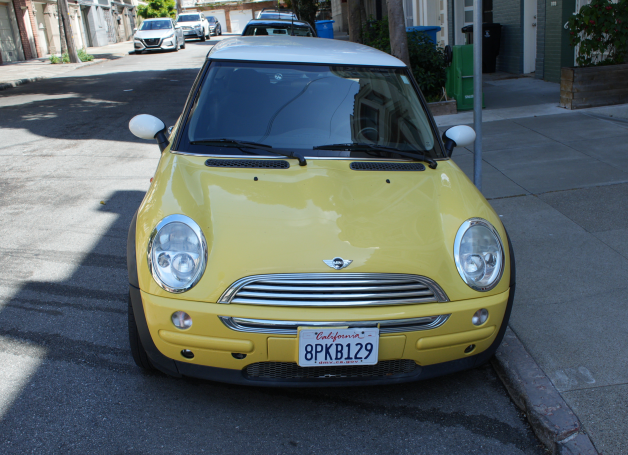} \\ 
        \rotatebox{90}{\hspace{10pt} \textbf{Hatchback}} &
        \includegraphics[width=0.19\linewidth]{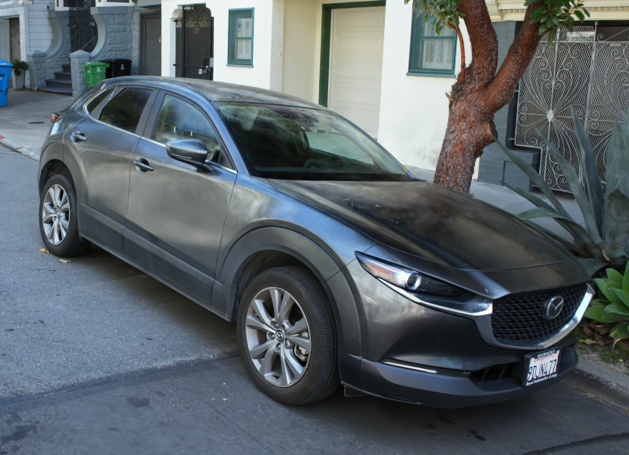} &
        \includegraphics[width=0.19\linewidth]{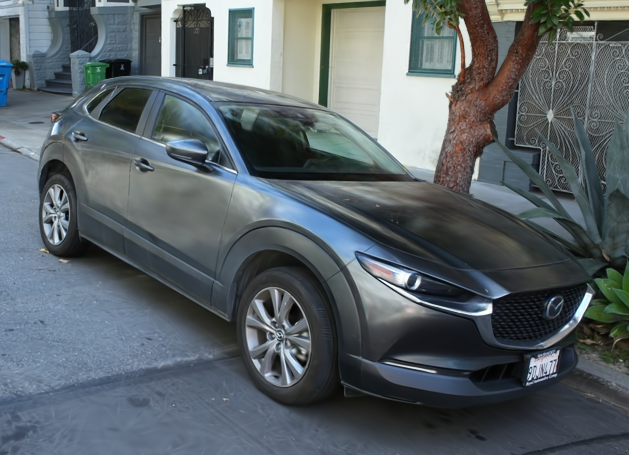} &
        \includegraphics[width=0.19\linewidth]{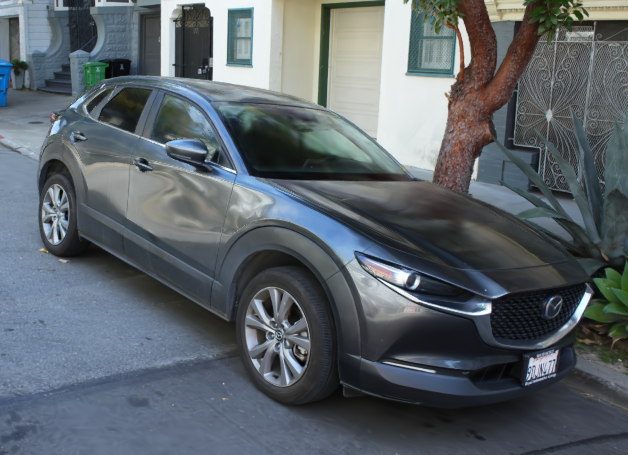} & 
        \includegraphics[width=0.19\linewidth]{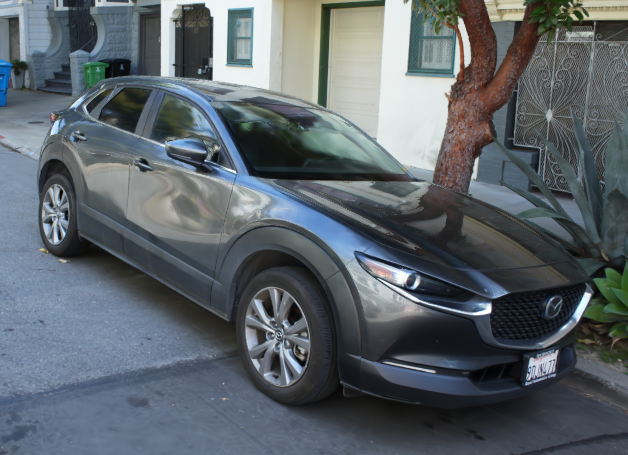} & 
        \includegraphics[width=0.19\linewidth]{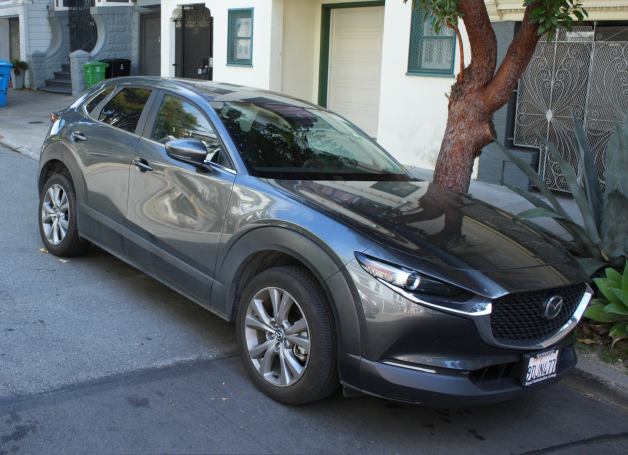} \\
        \rotatebox{90}{\hspace{20pt} \textbf{Toaster}} &
        \includegraphics[width=0.19\linewidth]{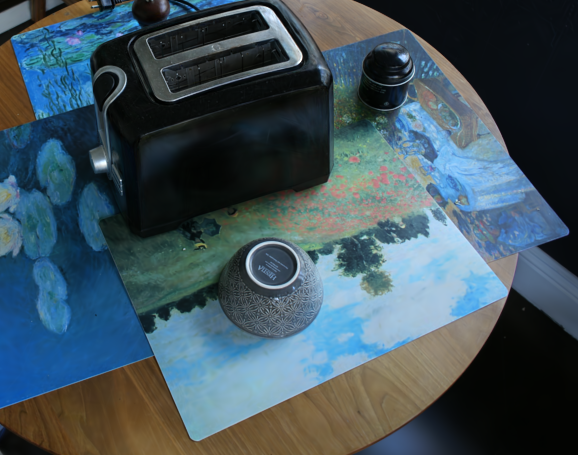} &
        \includegraphics[width=0.19\linewidth]{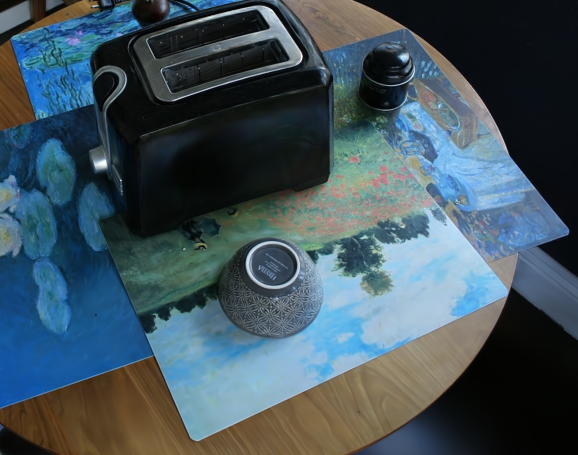} &
        \includegraphics[width=0.19\linewidth]{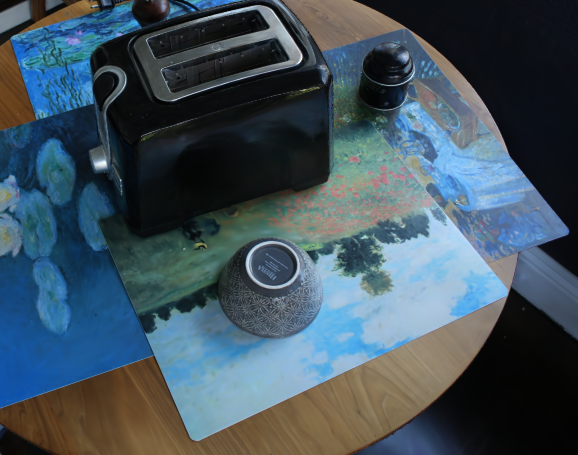} & 
        \includegraphics[width=0.19\linewidth]{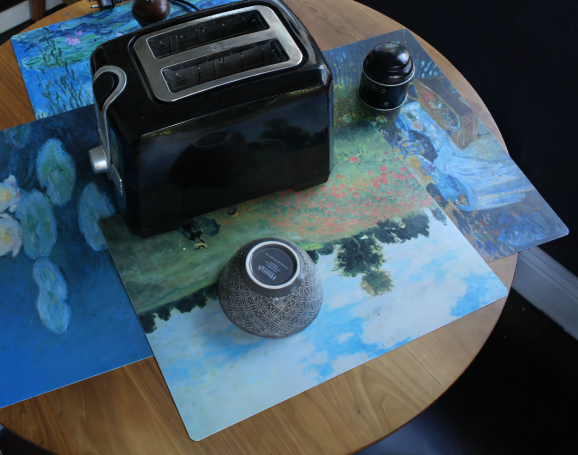} & 
        \includegraphics[width=0.19\linewidth]{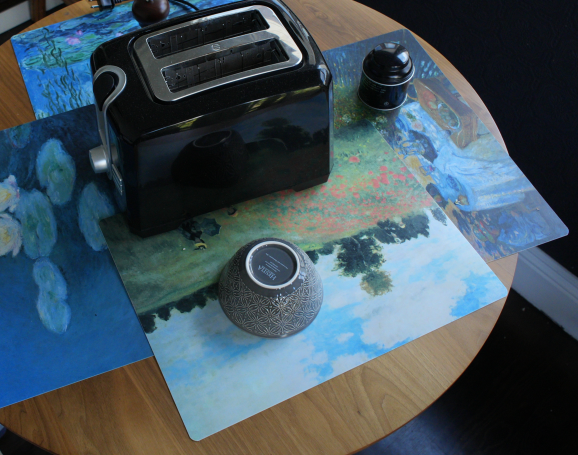} \\
        \rotatebox{90}{\hspace{40pt}\textbf{Dog}} &
        \includegraphics[width=0.19\linewidth]{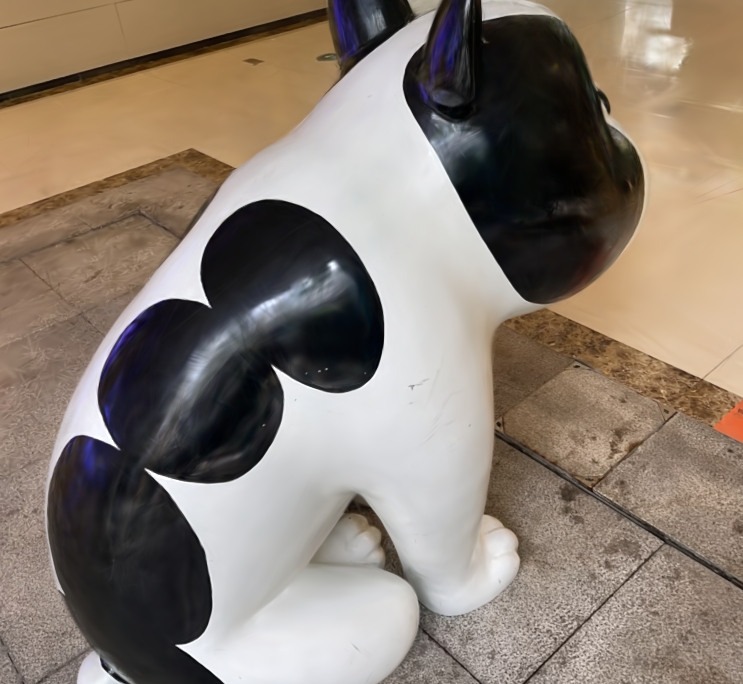} &
        \includegraphics[width=0.19\linewidth]{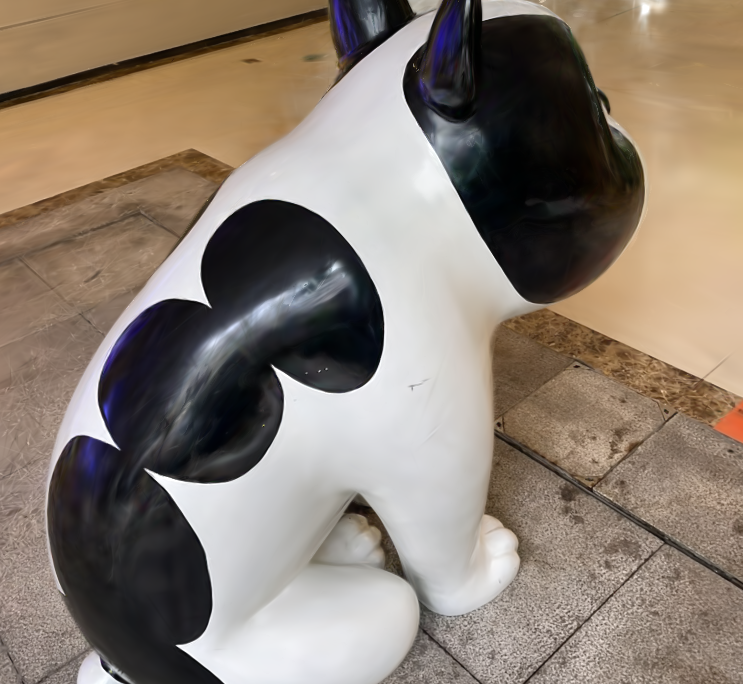} &
        \includegraphics[width=0.19\linewidth]{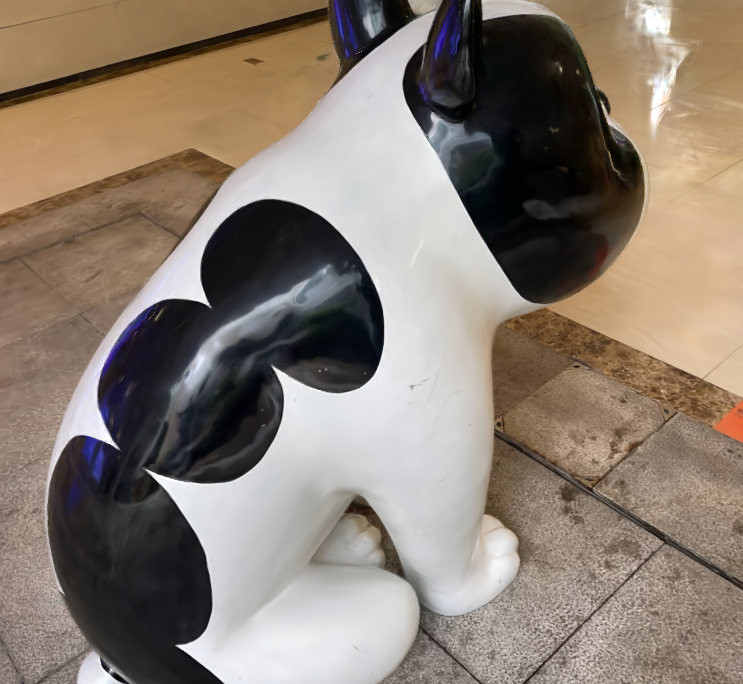} & 
        \includegraphics[width=0.19\linewidth]{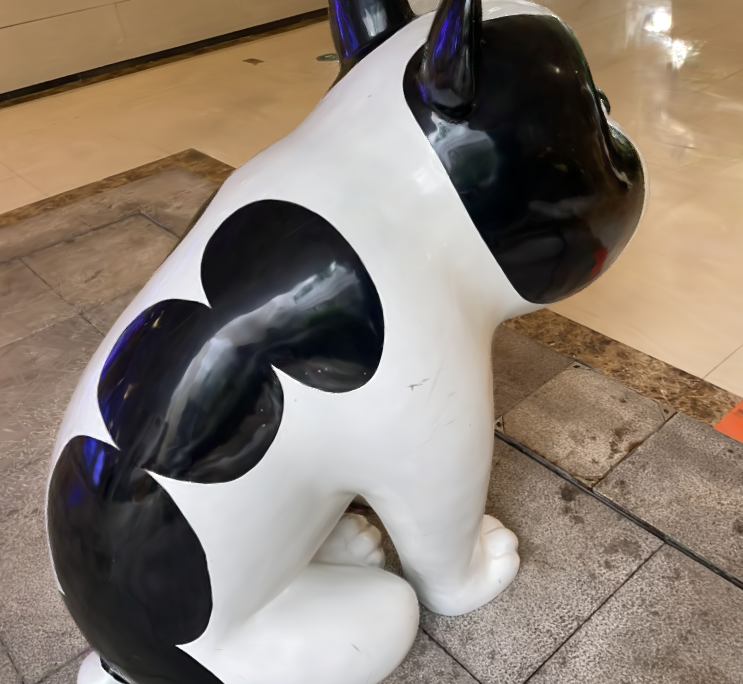} & 
        \includegraphics[width=0.19\linewidth]{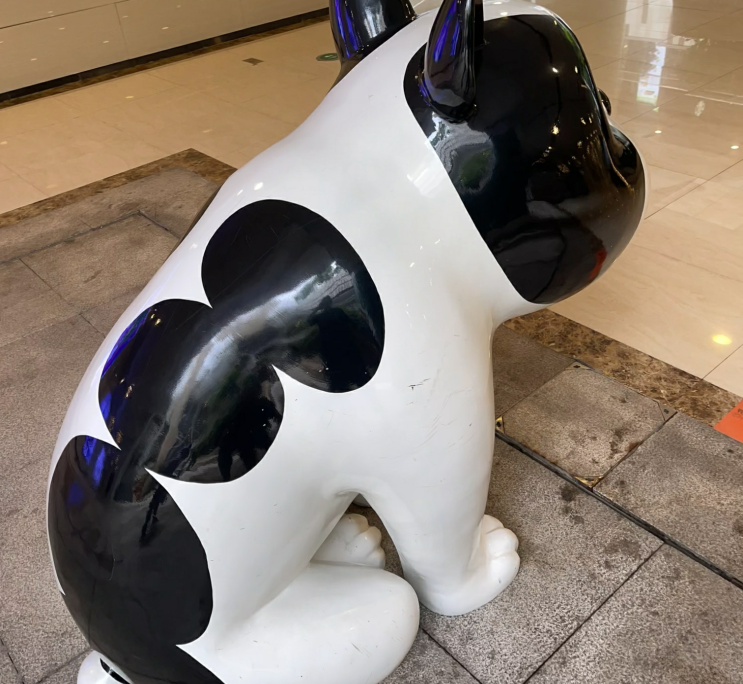} \\
        \rotatebox{90}{\hspace{17pt}\textbf{Garden}} &
        \includegraphics[width=0.19\linewidth]{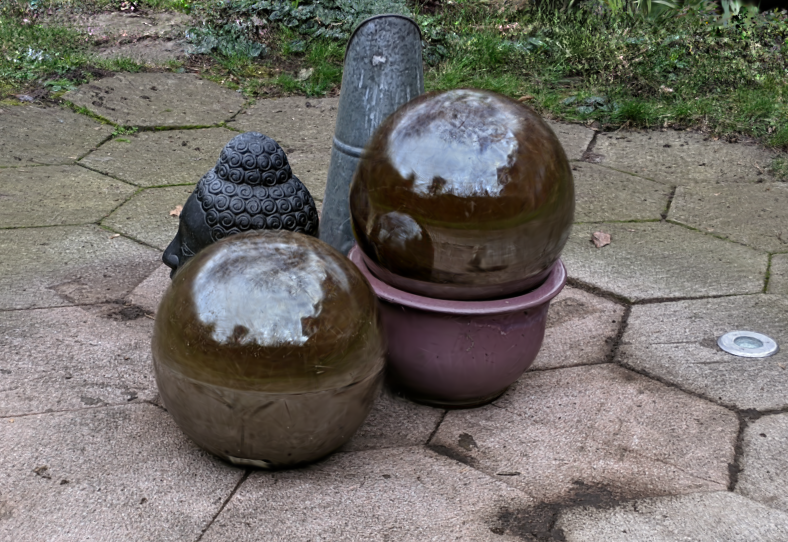} &
        \includegraphics[width=0.19\linewidth]{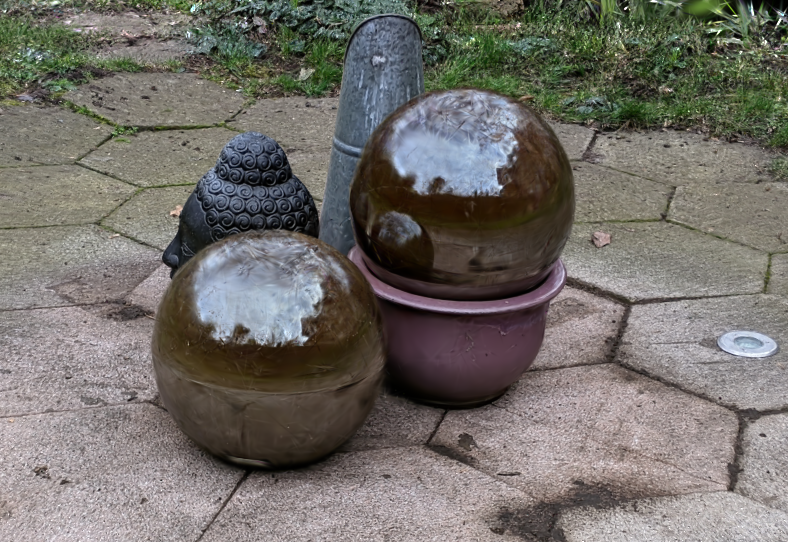} &
        \includegraphics[width=0.19\linewidth]{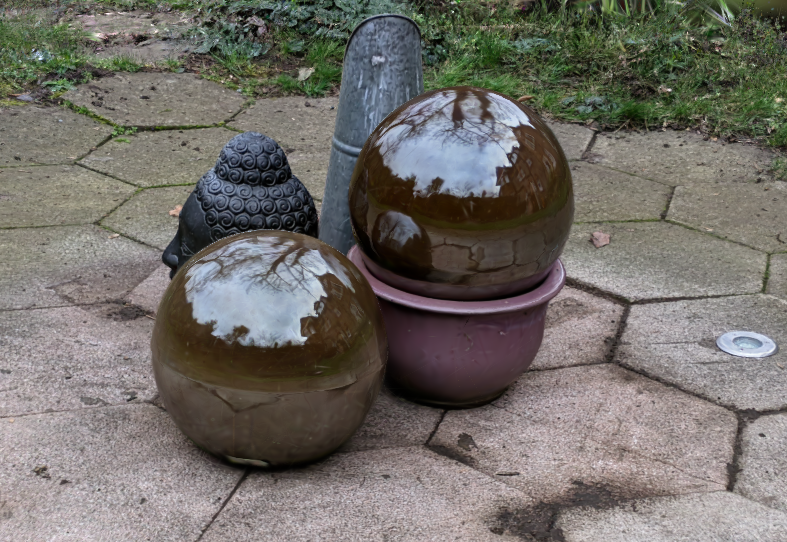} & 
        \includegraphics[width=0.19\linewidth]{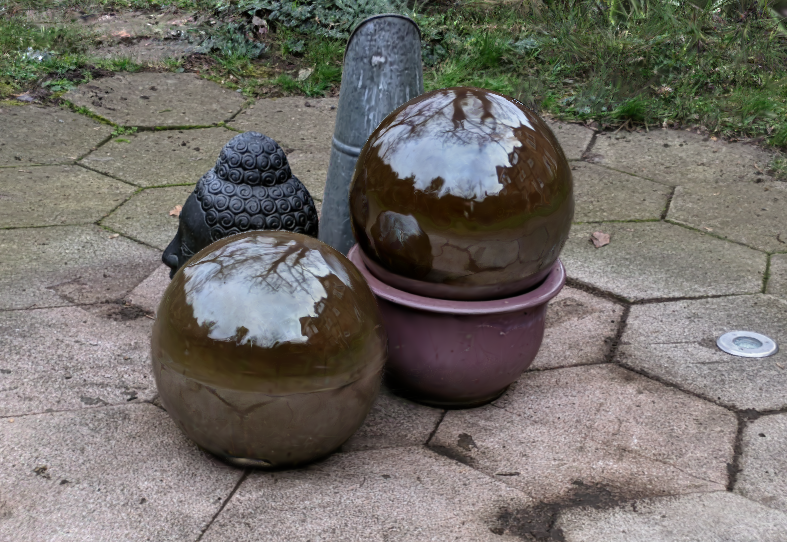} & 
        \includegraphics[width=0.19\linewidth]{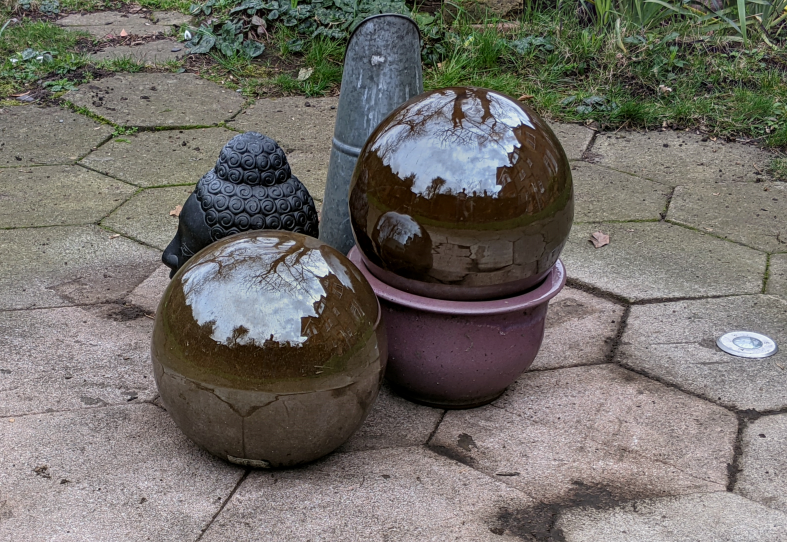} \\
        
        & 3DGS & 3DGS-DR & EnvGS & Ours & GT \\
    \end{tabular}
}
\caption{Additional qualitative comparison of novel view synthesis on the existing dataset.}
\label{fig:existing_dataset_nvs_supp}
\end{figure*}

We report per-scene quantitative results for novel view synthesis on the NeRF-Casting~\cite{nerfcasting}, EnvGS~\cite{envgs}, and Ref-Real~\cite{refnerf} datasets in Tab.~\ref{tab:existing_dataset_nvs_perscene}. Additional qualitative comparisons on the existing datasets and our custom NearRef dataset are shown in Fig.~\ref{fig:existing_dataset_nvs_supp} and Fig.~\ref{fig:custom_dataset_nvs_supp}, respectively.

\section{Custom NearRef Dataset}
To address the lack of datasets for evaluating novel view synthesis with near-field reflections, we collected a new dataset named NearRef. The dataset consists of four indoor scenes containing a large amount of near-field reflections: \textit{Black Monitor}, \textit{White Monitor}, \textit{Coffee Machine}, and \textit{Cleaner}.

All scenes are captured using the main camera of an iPhone 14 Pro. Each scene contains a highly reflective object surrounded by nearby items that produce near-field reflections. Videos are recorded at a resolution of $1080\times1920$ and $60$ FPS. The camera is moved around the object with a $180\degree - 360\degree$ horizontal arc with vertical motion.

\begin{figure}[t]
  \centering
  \scalebox{0.95}{
    \setlength{\tabcolsep}{1pt}
    \begin{tabular}{cc}
      \includegraphics[width=0.5\linewidth]{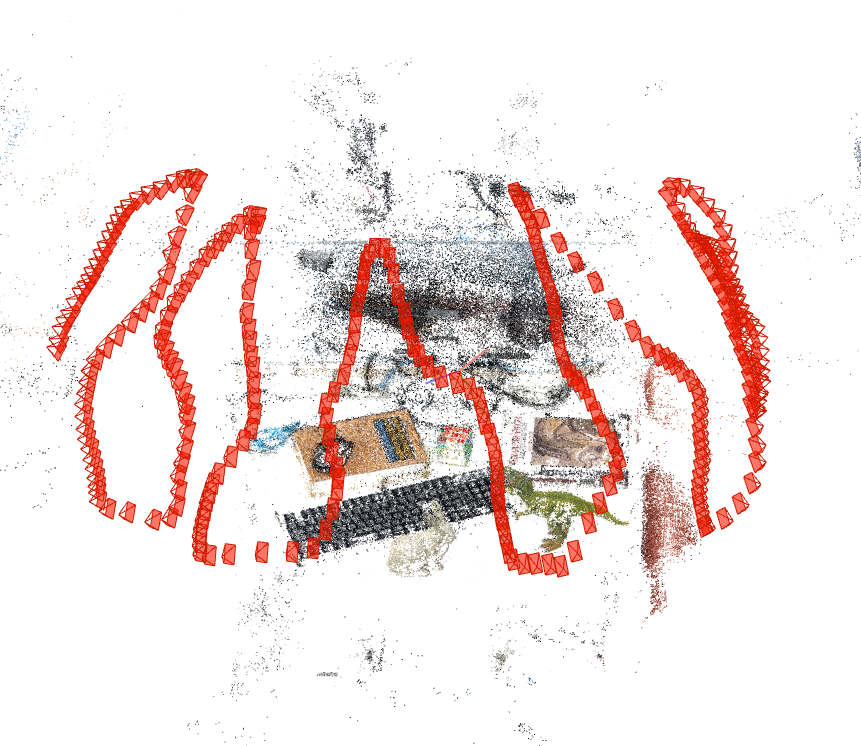} &
      \includegraphics[width=0.5\linewidth]{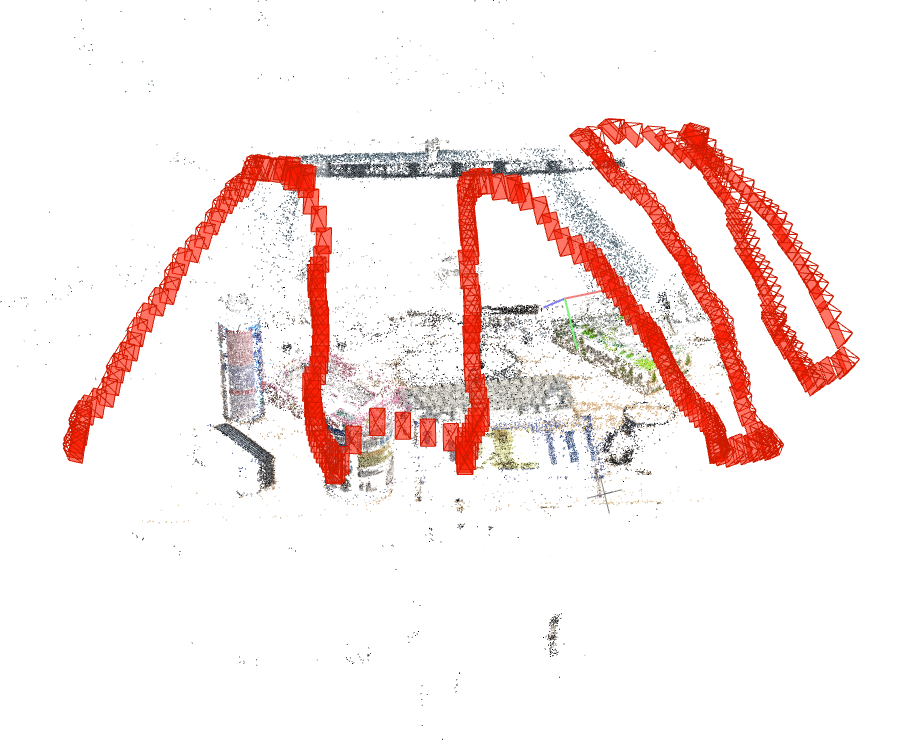} \\
      (a) Black Monitor & (b) White Monitor \\
      \includegraphics[width=0.5\linewidth]{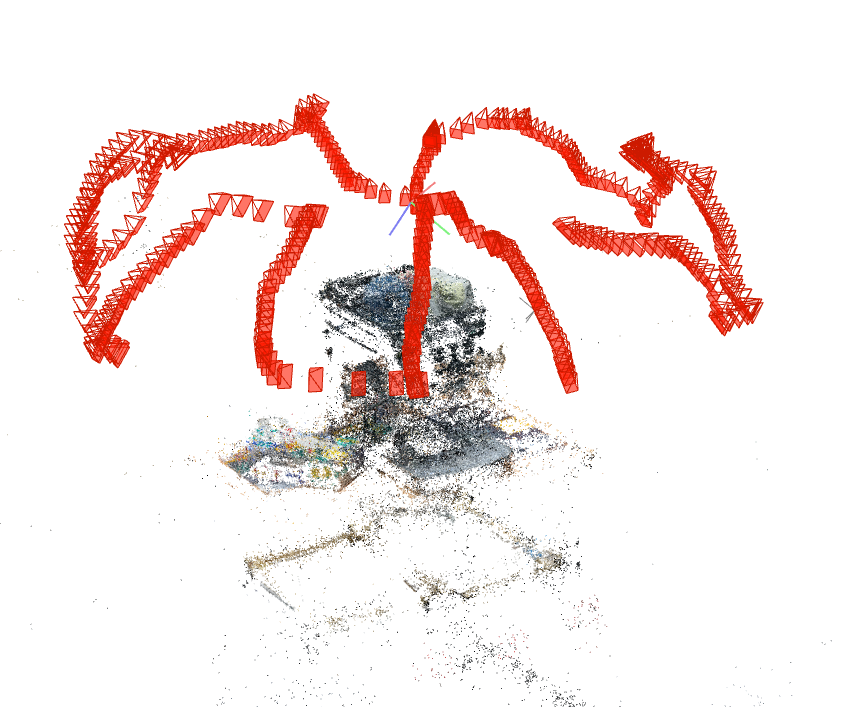} &
      \includegraphics[width=0.5\linewidth]{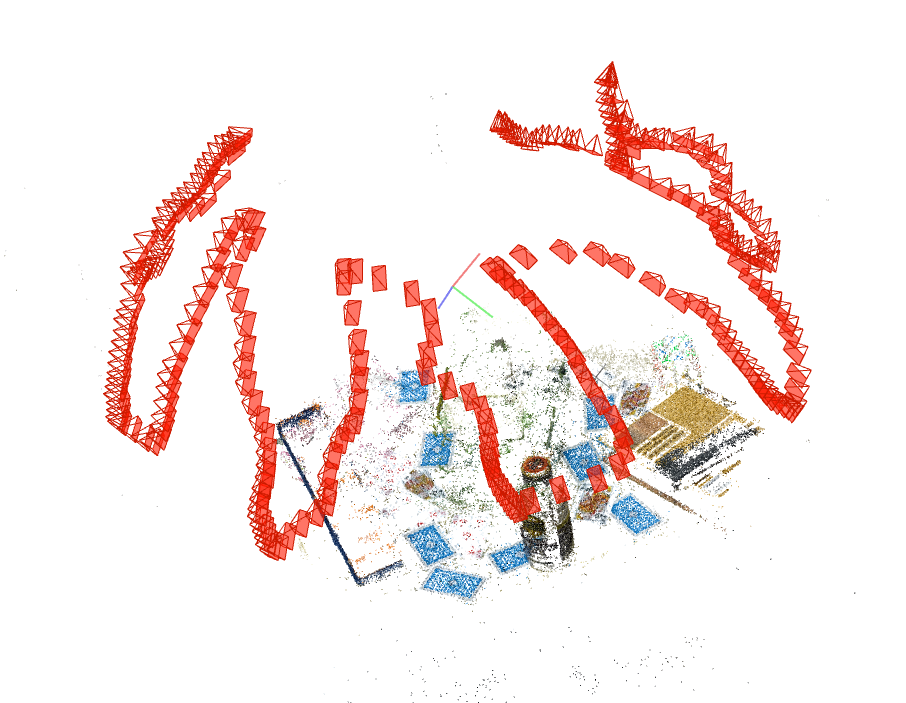} \\
      (c) Coffee Machine & (d) Cleaner \\
    \end{tabular}
  }
  \caption{Point cloud and camera trajectory of each scene in custom NearRef dataset.}
  \label{fig:colmap_points}
\end{figure}

The recorded videos are processed as follows. Frames without noticeable motion blur are uniformly sampled from each video and resized to $540\times960$. We use COLMAP~\cite{colmap} to estimate camera intrinsics, extrinsics, and a sparse 3D point cloud. The sparse point clouds and camera trajectories are visualized in Fig.~\ref{fig:colmap_points}.
All images are then undistorted using the estimated intrinsics and converted to a pinhole camera model to be compatible with the 2DGS~\cite{2dgs} framework.
For each sequence, every 8th frame is used for evaluation, while the remaining frames are used as input images. 

\section{Additional implementation details}
\subsection{Ray-Tracer}
We adopt the differentiable 2D Gaussian ray-tracer from EnvGS~\cite{envgs}. During rendering, each 2D Gaussian is approximated by two triangles that serve as proxy geometry. These triangles are inserted into a bounding volume hierarchy (BVH) for efficient intersection computation. Ray--Gaussian intersections are determined by traversing the BVH, evaluating the Gaussian response at each hit. The traversal continues until no further intersections are found or the accumulated transmittance falls below a predefined threshold.

\subsection{Optimization}
We provide the learning rates for per-Gaussian parameters as follows. For both primary and secondary Gaussians, the learning rates for the center, opacity, scale, and rotation are set to $1.6 \times 10^{-4}$, $5 \times 10^{-2}$, $5 \times 10^{-3}$, $1 \times 10^{-3}$, respectively. For both primary color and secondary color, we set the learning rate to $2.5 \times 10^{-3}$ for the DC component and $1.25 \times 10^{-4}$ for the higher-order components. The learning rate of specular weight $\bm{k}$ is set to $1\times10^{-2}$. The triplane and the MLP decoder are optimized with a learning rate of $1\times10^{-3}$. We optimize all parameters using the Adam optimizer, and adopt the densification and pruning strategies used in 2DGS~\cite{2dgs}.

\begin{table*}[ht]
\centering
\caption{Per-scene quantitative results of NVS on existing datasets~\cite{nerfcasting, envgs, refnerf}. \colorbox{red}{Best}, \colorbox{orange}{second best}, and \colorbox{yellow}{third best} results are highlighted.}
\renewcommand{\arraystretch}{0.9}
\scalebox{0.70}{ 
\begin{tabular}{l | ccc | ccc | ccc | ccc | ccc}
    \toprule
    & \multicolumn{12}{c|}{NeRF-Casting~\cite{nerfcasting}} & \multicolumn{3}{c}{EnvGS~\cite{envgs}}\\
    
    & \multicolumn{3}{c|}{Compact} & \multicolumn{3}{c|}{Grinder} & \multicolumn{3}{c|}{Hatchback} & \multicolumn{3}{c|}{Toaster} & \multicolumn{3}{c}{Audi} \\
    Method & PSNR$\uparrow$ & SSIM$\uparrow$ & LPIPS$\downarrow$ & PSNR$\uparrow$ & SSIM$\uparrow$ & LPIPS$\downarrow$ & PSNR$\uparrow$ & SSIM$\uparrow$ & LPIPS$\downarrow$ & PSNR$\uparrow$ & SSIM$\uparrow$ & LPIPS$\downarrow$ & PSNR$\uparrow$ & SSIM$\uparrow$ & LPIPS$\downarrow$ \\
    \midrule
    3DGS~\cite{3dgs} & \cellcolor{red}30.1704 & \cellcolor{red}0.9090 & \cellcolor{red}0.1371 & \cellcolor{yellow}31.6084 & \cellcolor{orange}0.8598 & \cellcolor{red}0.1696 & \cellcolor{yellow}27.1412 & \cellcolor{red}0.8703 & \cellcolor{red}0.1567 & \cellcolor{yellow}29.9444 & \cellcolor{yellow}0.9328 & \cellcolor{yellow}0.1133 & \cellcolor{yellow}27.9984 & \cellcolor{red}0.8849 & \cellcolor{orange}0.1296 \\
    2DGS~\cite{2dgs} & 29.6057 & \cellcolor{orange}0.8880 & 0.1758 & 29.9115 & 0.8374 & 0.2165 & 26.5643 & \cellcolor{yellow}0.8472 & 0.1983 & 28.9365 & 0.9186 & 0.1395 & 27.5342 & 0.8750 & 0.1628 \\
    GShader~\cite{gaussianshader} & 28.2558 & 0.8692 & 0.2005 & 26.9084 & 0.7899 & 0.3108 & 25.7765 & 0.8248 & 0.2222 & 27.6720 & 0.8972 & 0.1638 & 26.5692 & 0.8503 & 0.1942 \\
    3DGS-DR~\cite{3dgsdr} & 29.7766 & \cellcolor{yellow}0.8848 & 0.1851 & 29.8010 & \cellcolor{red}0.8766 & 0.2198 & 26.5551 & 0.8358 & 0.2222 & 29.4645 & 0.9248 & 0.1554 & 27.8997 & 0.8735 & 0.1644 \\
    RefGS~\cite{refgs} & 25.2219 & 0.8336 & 0.2243 & 30.0620 & 0.8272 & 0.2443 & 18.6603 & 0.7401 & 0.2831 & 23.0695 & 0.8478 & 0.2117 & 27.5613 & 0.8572 & 0.1952 \\
    EnvGS~\cite{envgs} & \cellcolor{yellow}29.6681 & 0.8736 & \cellcolor{yellow}0.1660 & \cellcolor{orange}32.4982 & 0.8428 & \cellcolor{orange}0.1712 & \cellcolor{orange}27.4097 & 0.8437 & \cellcolor{yellow}0.1827 & \cellcolor{orange}31.7538 & \cellcolor{orange}0.9334 & \cellcolor{orange}0.1011 & \cellcolor{orange}28.8483 & \cellcolor{yellow}0.8812 & \cellcolor{yellow}0.1397 \\
    \textbf{Ours} & \cellcolor{orange}30.0102 & 0.8769 & \cellcolor{orange}0.1613 & \cellcolor{red}32.5457 & \cellcolor{yellow}0.8438 & \cellcolor{yellow}0.1714 & \cellcolor{red}27.7905 & \cellcolor{orange}0.8497 & \cellcolor{orange}0.1752 & \cellcolor{red}32.1275 & \cellcolor{red}0.9344 & \cellcolor{red}0.1011 & \cellcolor{red}29.0587 & \cellcolor{orange}0.8844 & \cellcolor{red}0.1343 \\
    \midrule
    & \multicolumn{3}{c|}{EnvGS~\cite{envgs}} & \multicolumn{9}{c|}{Ref-Real~\cite{refnerf}} & &&\\
    & \multicolumn{3}{c|}{Dog} & \multicolumn{3}{c|}{Garden} & \multicolumn{3}{c|}{Sedan} & \multicolumn{3}{c|}{Toycar} & \multicolumn{3}{c}{\textbf{Average}} \\
    Method & PSNR$\uparrow$ & SSIM$\uparrow$ & LPIPS$\downarrow$ & PSNR$\uparrow$ & SSIM$\uparrow$ & LPIPS$\downarrow$ & PSNR$\uparrow$ & SSIM$\uparrow$ & LPIPS$\downarrow$ & PSNR$\uparrow$ & SSIM$\uparrow$ & LPIPS$\downarrow$ & PSNR$\uparrow$ & SSIM$\uparrow$ & LPIPS$\downarrow$ \\
    \midrule
    3DGS~\cite{3dgs} & \cellcolor{yellow}25.2696 & \cellcolor{orange}0.8814 & \cellcolor{yellow}0.1654 & 22.3905 & 0.5934 & \cellcolor{yellow}0.2378 & \cellcolor{yellow}25.4761 & \cellcolor{red}0.7280 & \cellcolor{red}0.2943 & \cellcolor{yellow}24.3373 & 0.6587 & \cellcolor{red}0.2275 & \cellcolor{yellow}27.1485 & \cellcolor{yellow}0.8131 & \cellcolor{red}0.1824 \\
    2DGS~\cite{2dgs} &24.9589&0.8759&0.1849& \cellcolor{yellow}22.5432 & \cellcolor{yellow}0.6079 & 0.2548 & 25.2721 & \cellcolor{yellow}0.7191 & 0.3320 & 23.8158 & \cellcolor{red}0.6287 & 0.2379 & 26.5714 & 0.8057 & 0.2214  \\
    GShader~\cite{gaussianshader} &24.3289&0.8604&0.2022& 22.0699 & 0.5979 & 0.2691 & 23.8579 & 0.6750 & 0.3677 & 23.3839 & 0.6385 & 0.2928 & 25.4247 & 0.7781 & 0.2470  \\
    3DGS-DR~\cite{3dgsdr} & 23.8829 & 0.8453 & 0.2172 & 21.5413 & 0.5707 & 0.2494 & 23.9719 & 0.6932 & 0.3398 & 23.1681 & 0.6270 & 0.2848 & 26.2290 & 0.7924 & 0.2264 \\
    RefGS~\cite{refgs} & 23.7392 & 0.8500 & 0.2077 & 22.2284 & 0.5879 & 0.2748 & 25.0043 & 0.7021 & 0.3367 & 24.2843 & 0.6581 & 0.2577 & 24.4257 & 0.7671 & 0.2484\\
    EnvGS~\cite{envgs} & \cellcolor{orange}25.6849 & \cellcolor{yellow}0.8806 & \cellcolor{orange}0.1551 & \cellcolor{red}23.7025 & \cellcolor{red}0.6192 & \cellcolor{red}0.1998 & \cellcolor{red}26.0768 & \cellcolor{orange}0.7256 & \cellcolor{orange}0.3002 & \cellcolor{orange}24.6423 & \cellcolor{yellow}0.6671 & \cellcolor{orange}0.2315 & \cellcolor{orange}27.7394 & \cellcolor{orange}0.8075 & \cellcolor{yellow}0.1834 \\
    \textbf{Ours} & \cellcolor{red}25.7860 & \cellcolor{red}0.8826 & \cellcolor{red}0.1540 & \cellcolor{orange}23.0693 & \cellcolor{orange}0.6188 & \cellcolor{orange}0.2004 & \cellcolor{orange}25.5214 & 0.7173 & \cellcolor{yellow}0.3074 & \cellcolor{red}24.6526 & \cellcolor{orange}0.6684 & \cellcolor{yellow}0.2320 & \cellcolor{red}27.8402 & \cellcolor{red}0.8085 & \cellcolor{red}0.1819 \\
    \bottomrule
\end{tabular}}
\label{tab:existing_dataset_nvs_perscene}
\end{table*}
\subsection{Losses}
Our overall training loss shown in Eq.7 in the main paper consists of three terms: an image reconstruction loss $\mathcal{L}_{\text{rgb}}$, a normal consistency loss $\mathcal{L}_{\text{consist}}$, and a monocular normal loss $\mathcal{L}_{\text{mono}}$. The image reconstruction loss minimizes photometric difference between the rendered image and the ground truth image. It is defined as a weighted combination of $\mathcal{L}_1$ loss, D-SSIM~\cite{SSIM} loss, and LPIPS~\cite{lpips} loss:
\begin{equation}
    \mathcal{L}_{\text{rgb}} = \lambda_{\text{L1}}\mathcal{L}_{\text{L1}}+\lambda_{\text{D-SSIM}}\mathcal{L}_{\text{D-SSIM}} + \lambda_{\text{LPIPS}}\mathcal{L}_{\text{LPIPS}},
\end{equation} 
where the weights $\lambda_{\text{L1}}$, $\lambda_{\text{D-SSIM}}$, $\lambda_{\text{LPIPS}}$ are set to 0.8, 0.2 and 0.01, respectively. 

Following 2DGS~\cite{2dgs}, we add a normal consistency loss $\mathcal{L}_{\text{consist}}$ to enforce the alignment between the rendered normal and the normal derived from the depth map:
\begin{equation}
    \mathcal{L}_{\text{consist}} = \frac{1}{N_p}\sum_i^{N_p} (1 - \bm{N}_i\cdot\bm{N}_{\text{depth},i}),
\end{equation}
where $N_p$ is the number of image pixels. 
$\bm{N}_{\text{depth}}$ is derived from the gradient of rendered depth map, while $\bm{N}_i$ denotes the rendered normal. The weight of $\mathcal{L}_{\text{consist}}$ is set to $0.04$.

We further supervise the rendered normal using monocular normal estimates $\bm{N}_{\text{mono}}$ from a pretrained monocular normal estimation model as:
\begin{equation}
    \mathcal{L}_{\text{mono}} = \frac{1}{N_p} \sum_i^{N_p} (1 - \bm{N}_i\cdot \bm{N}_{\text{mono}}).
\end{equation}
The weight of $\mathcal{L}_{\text{mono}}$ is set to $0.01$.

\begin{figure*}[t]
\centering
\setlength{\tabcolsep}{1pt}
\renewcommand{\arraystretch}{0.5}
\scalebox{0.9}{
    \begin{tabular}{@{}cccccc@{}}
        \rotatebox{90}{\hspace{40pt} \textbf{Black Monitor}} &
        \includegraphics[width=0.19\linewidth]{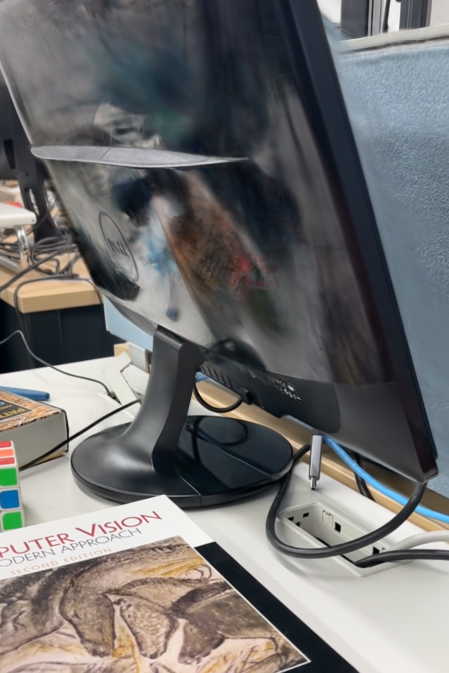} &
        \includegraphics[width=0.19\linewidth]{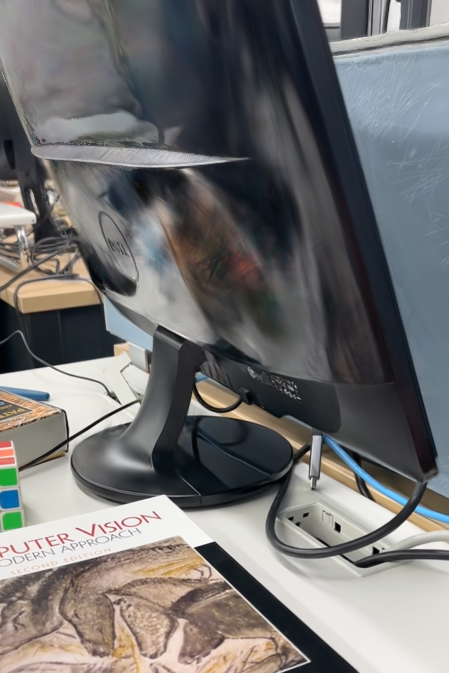} & 
        \includegraphics[width=0.19\linewidth]{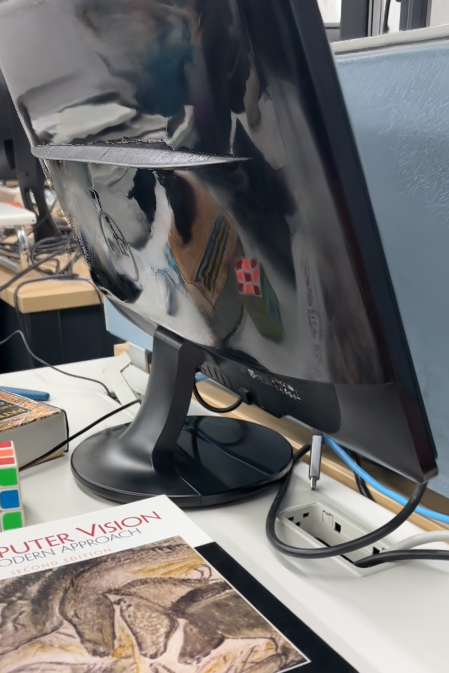} & 
        \includegraphics[width=0.19\linewidth]{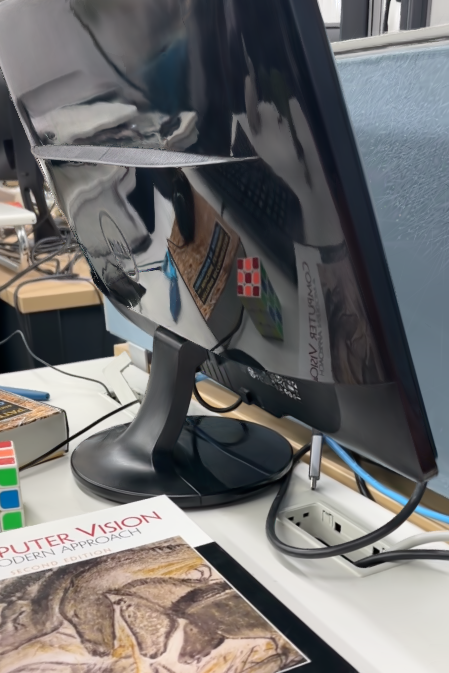} & 
        \includegraphics[width=0.19\linewidth]{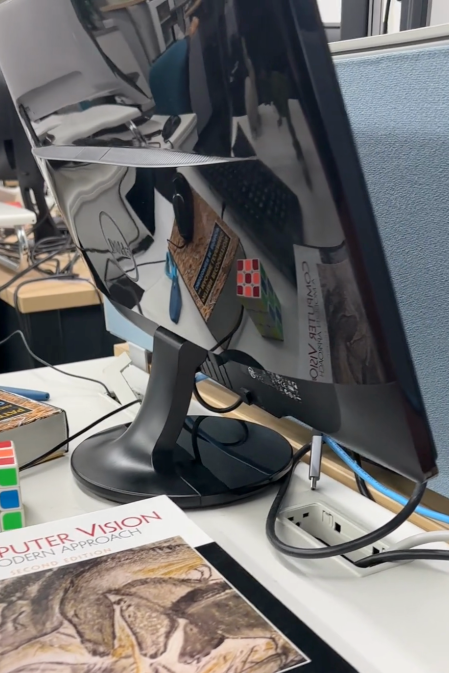} \\ 
        \rotatebox{90}{\hspace{40pt} \textbf{White Monitor}} &
        \includegraphics[width=0.19\linewidth]{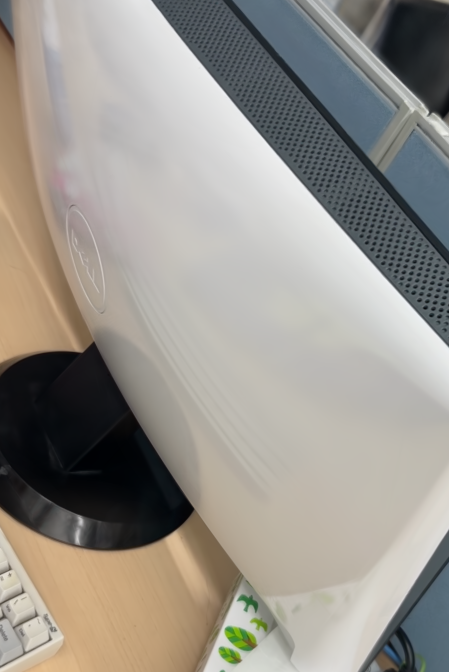} &
        \includegraphics[width=0.19\linewidth]{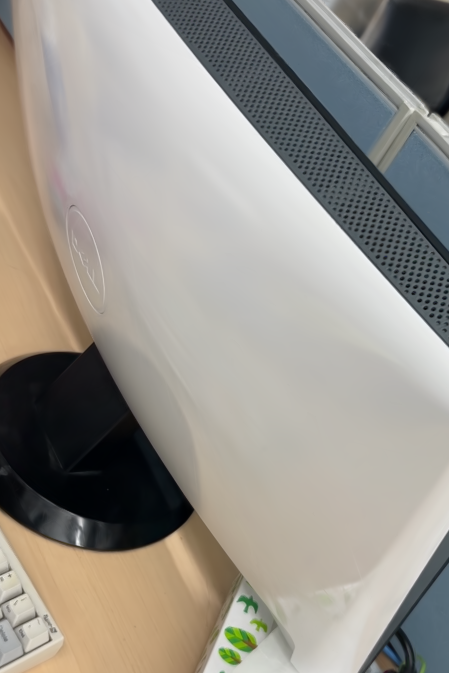} & 
        \includegraphics[width=0.19\linewidth]{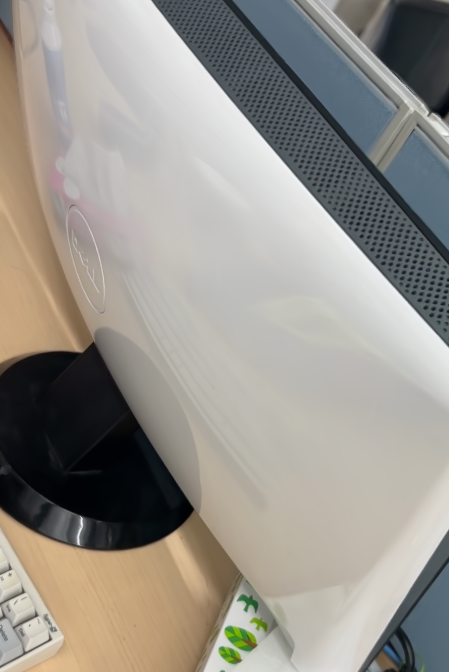} & 
        \includegraphics[width=0.19\linewidth]{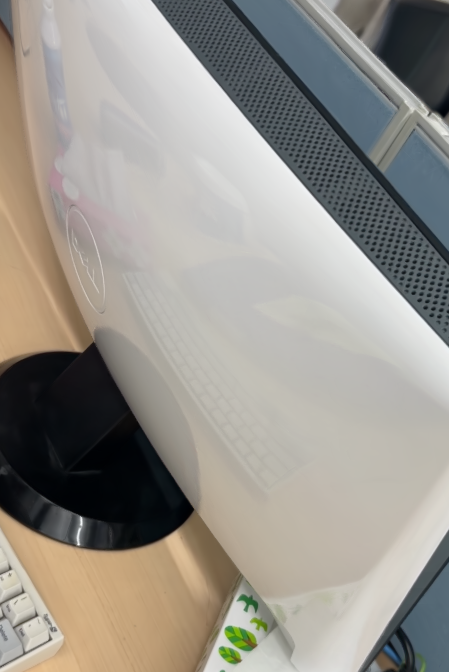} & 
        \includegraphics[width=0.19\linewidth]{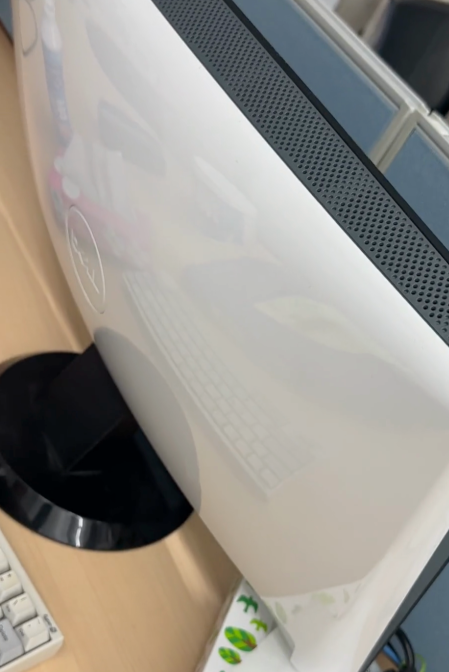} \\ 
        \rotatebox{90}{\hspace{40pt} \textbf{Coffee Machine}} &
        \includegraphics[width=0.19\linewidth]{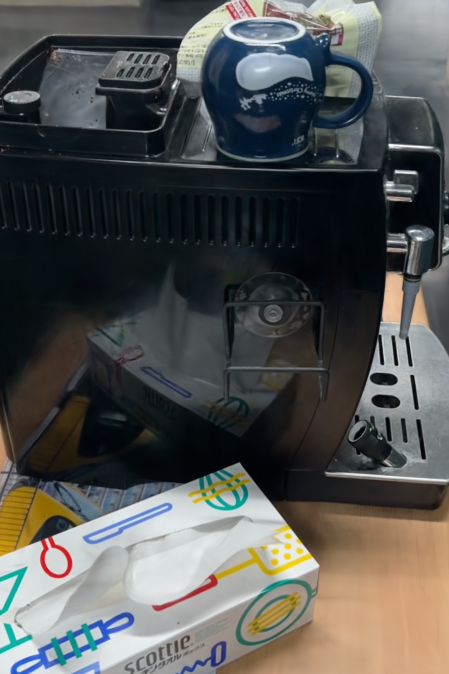} &
        \includegraphics[width=0.19\linewidth]{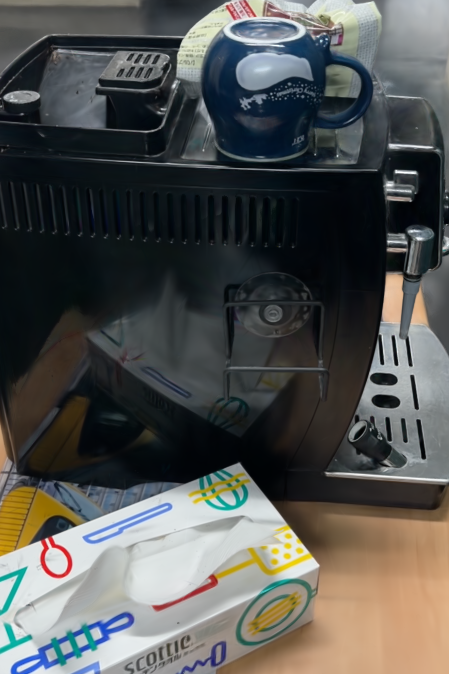} & 
        \includegraphics[width=0.19\linewidth]{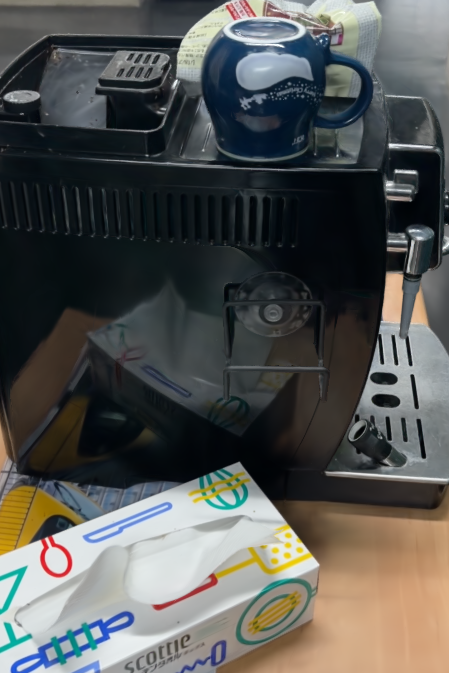} & 
        \includegraphics[width=0.19\linewidth]{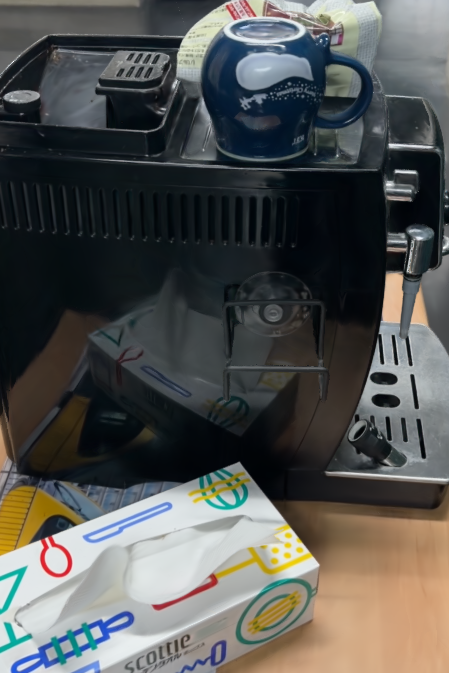} & 
        \includegraphics[width=0.19\linewidth]{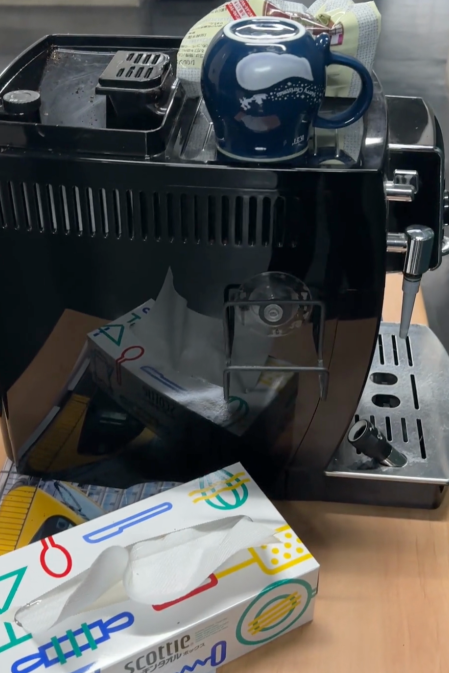} \\ 
        \rotatebox{90}{\hspace{55pt} \textbf{Cleaner}} &
        \includegraphics[width=0.19\linewidth]{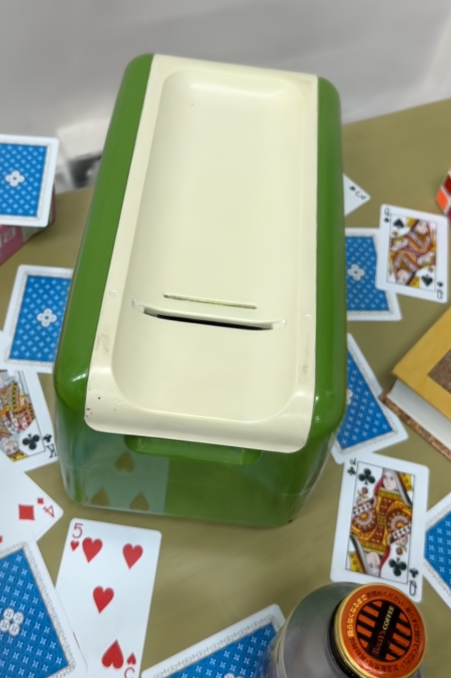} &
        \includegraphics[width=0.19\linewidth]{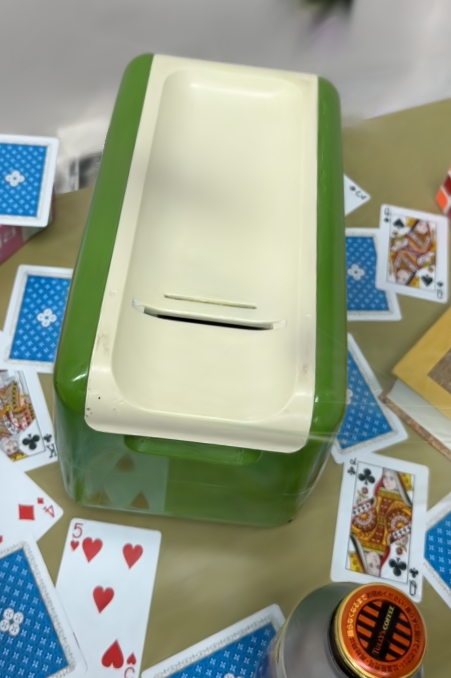} &
        \includegraphics[width=0.19\linewidth]{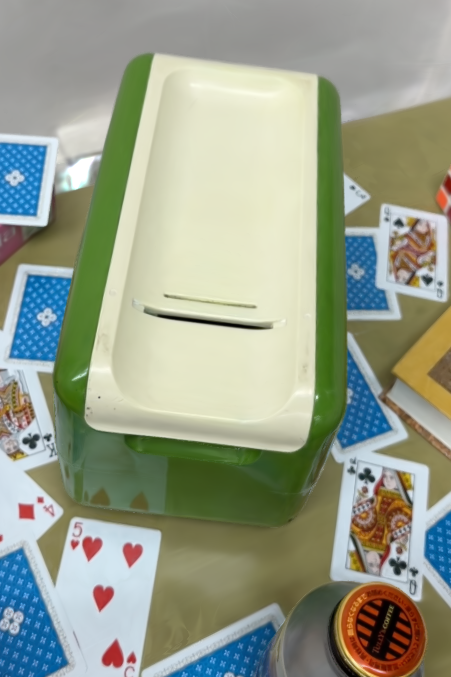} & 
        \includegraphics[width=0.19\linewidth]{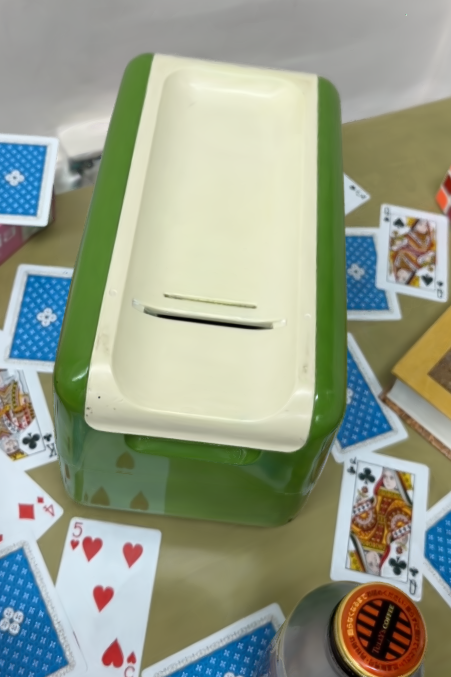} & 
        \includegraphics[width=0.19\linewidth]{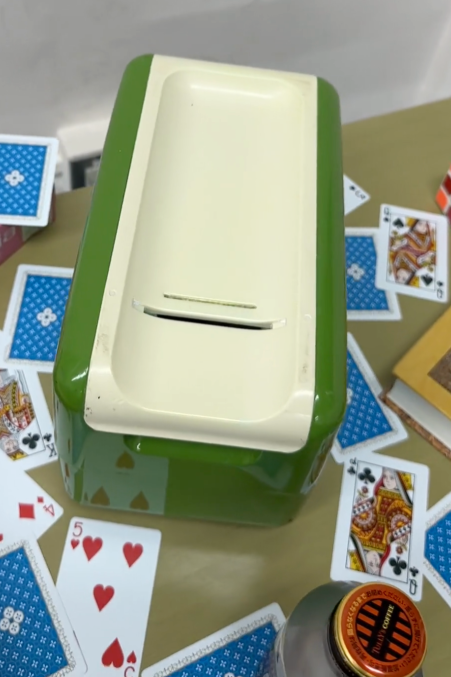} \\
        
        & 3DGS & 3DGS-DR & EnvGS & Ours & GT \\
    \end{tabular}
}
\caption{Additional qualitative comparison of NVS on custom NearRef dataset.}
\label{fig:custom_dataset_nvs_supp}
\end{figure*}

\end{document}